\documentclass{article}

\usepackage{PRIMEarxiv}

\usepackage[utf8]{inputenc} 
\usepackage[T1]{fontenc}    
\usepackage{hyperref}       
\usepackage{url}            
\usepackage{booktabs}       
\usepackage{amsfonts}       
\usepackage{nicefrac}       
\usepackage{microtype}      
\usepackage{lipsum}
\usepackage{fancyhdr}       
\usepackage{graphicx}       

\usepackage{cite}
\usepackage{amsmath,amssymb}
\usepackage{algorithmic}
\usepackage{textcomp}
\usepackage{xcolor}

\usepackage{subcaption}
\usepackage[multiple]{footmisc}
\usepackage{dsfont}
\DeclareMathOperator*{\argmax}{arg\,max}

\newcounter{definition}
\newtheorem{definition}{Definition}

\title{Investigating and unmasking feature-level vulnerabilities of CNNs to adversarial perturbations
}

\author{
  Davide Coppola\\
  Bioinformatics Institute (BII), Agency for Science, Technology and Research (A*STAR), Singapore\\
  \texttt{davidec@bii.a-star.edu.sg} \\
  \And
  Hwee Kuan Lee \\
  Bioinformatics Institute (BII), Agency for Science, Technology and Research (A*STAR), Singapore \\
  School of Computing, National University of Singapore, Singapore \\
  School of Biological Sciences, Nanyang Technological University, Singapore \\
  Singapore Eye Research Institute (SERI), Singapore \\
  International Research Laboratory on Artificial Intelligence, Singapore \\
  Singapore Institute for Clinical Sciences, Singapore \\
  \texttt{leehk@bii.a-star.edu.sg}
   \\
}

\begin{document}
\maketitle

\begin{abstract}
This study explores the impact of adversarial perturbations on Convolutional Neural Networks (CNNs) with the aim of enhancing the understanding of their underlying mechanisms. Despite numerous defense methods proposed in the literature, there is still an incomplete understanding of this phenomenon.
Instead of treating the entire model as vulnerable, we propose that specific feature maps learned during training contribute to the overall vulnerability. To investigate how the hidden representations learned by a CNN affect its vulnerability, we introduce the \textit{Adversarial Intervention} framework.
Experiments were conducted on models trained on three well-known computer vision datasets, subjecting them to attacks of different nature. Our focus centers on the effects that adversarial perturbations to a model's initial layer have on the overall behavior of the model.
Empirical results revealed compelling insights: a) perturbing selected channel combinations in shallow layers causes significant disruptions; b) the channel combinations most responsible for the disruptions are common among different types of attacks; c) despite shared vulnerable combinations of channels, different attacks affect hidden representations with varying magnitudes; d) there exists a positive correlation between a kernel's magnitude and its vulnerability. 
In conclusion, this work introduces a novel framework to study the vulnerability of a CNN model to adversarial perturbations, revealing insights that contribute to a deeper understanding of the phenomenon. 
The identified properties pave the way for the development of efficient ad-hoc defense mechanisms in future applications.
\end{abstract}


\section{Introduction}

The vulnerability of Deep Learning models to adversarial perturbations restricts their usability in high risk situations \cite{qiu2019ReviewArtificialIntelligence,wang2022AdversarialAttacksDefenses}. Despite the efforts that have been made in the literature to understand the causes that lead to the adversarial perturbations phenomenon \cite{szegedy2014IntriguingPropertiesNeural, goodfellow2015ExplainingHarnessingAdversarial, gilmer2019AdversarialExamplesAre, ilyas2019AdversarialExamplesAre, paknezhad2022ExplainingAdversarialVulnerability}, there is no consensus on the mechanism of how adversarial perturbations cause failures in the prediction of models with good performance.
This paper contributes to the understanding of the vulnerability of Convolutional Neural Network (CNN) models to adversarial perturbations from the point of view of the feature representations learned by a model after the training phase has completed. 
Several studies have looked into quantifying the vulnerability of features in individual channels in terms of their sensitivity to adversarial perturbations \cite{zhang2021AdversarialPerturbationDefense, bai2022ImprovingAdversarialRobustness, yan2021CIFSImprovingAdversarial, khalooei2022LayerwiseRegularizedAdversarial}. However, sensitivity analyses mainly display the correlations between channel activations and adversarial perturbations but do not provide information on how each channel causes the entire model to fail. This paper investigates the adversarial phenomenon by performing interventions on the activations of CNN channels.

\begin{figure}
	\centering
	\includegraphics[width=0.90\linewidth]{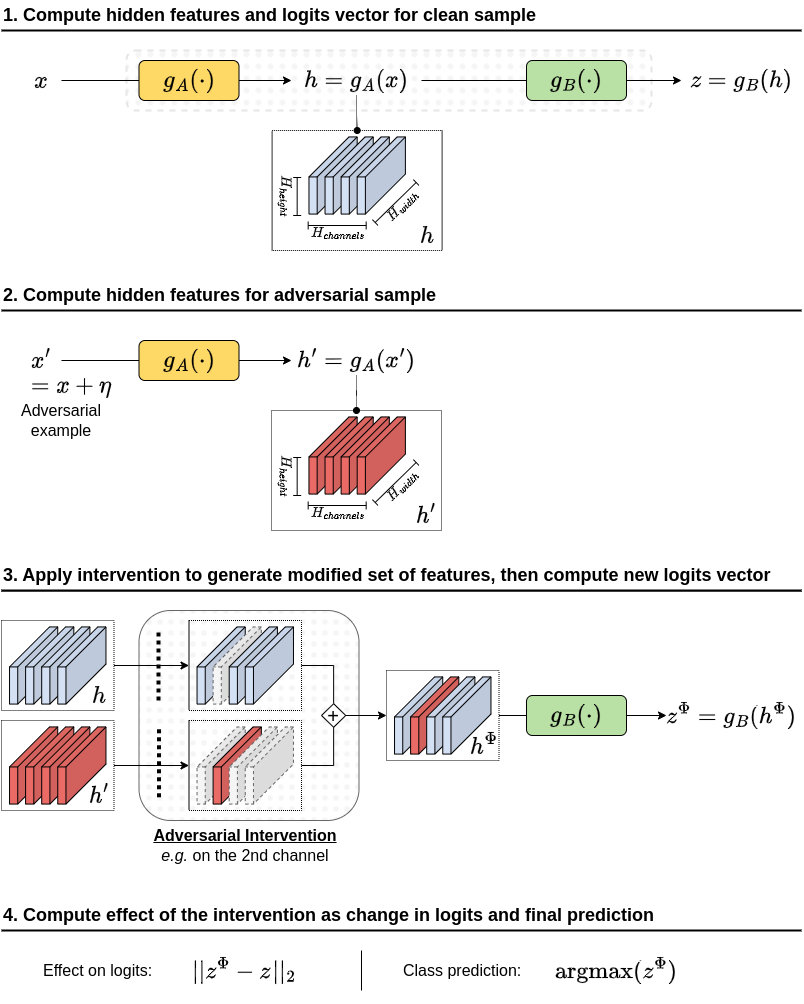}
	\caption{
		Schematic of the \textit{Adversarial Intervention} framework.
		A CNN model is given, which can be expressed as two consecutive functions $g_A$ and $g_B$.
		\textit{1.} Given a clean sample $x$, the intermediate feature representation $h=g_A(x)$ and the output logits $z=g_B(g_A(x))$ are computed.
		\textit{2.} The adversarial example $x'=x+\eta$ is computed using an arbitrary adversarial attack algorithm from the literature. Subsequently, the corrupted intermediate feature representation is computed $h'=g_A(x')$.
		\textit{3.} The \textit{Adversarial Intervention} operation consists in swapping $\gamma$ arbitrary channels in the the clean representation $h$ with their equivalents in the corrupted representation $h'$; this is achieved through a masking operation. 
		The resulting set of features $h^\Phi$ is then used to compute the output logits $z^\Phi=g_B(h^\Phi)$.
		\textit{4.} The effect of the intervention is evaluated by computing various metrics.
	}
	\label{fig:adversarial-intervention-framework}
\end{figure}

Following the intuition that small perturbations at the start of a model follow a “snowball effect” that increases their magnitude as they travel along the model architecture \cite{goodfellow2015ExplainingHarnessingAdversarial, zhang2021AdversarialPerturbationDefense}, we hypothesise and show that certain specific channels in the shallow layers of a CNN may be among the leading causes of a model’s failure to perform the classification correctly.
The \textit{Adversarial Intervention} framework (Figure \ref{fig:adversarial-intervention-framework}) has been developed in this work to provide evidence to the claim. It is established that to study cause-effect relationships, it is necessary to apply interventions on the system of interest \cite{pearl2009CausalityModelsReasoning,chattopadhyay2019NeuralNetworkAttributions}. Hence, the proposed framework consists of modifying the clean (unperturbed) hidden representations by substituting some of their channels with their corresponding adversarially perturbed representations. The changes in the output caused by the controlled substitution reveal the contributions of individual channels to prediction failures.

If the perturbation of a relatively small set of channels A results in frequent failures, whereas perturbing another equally small set B has a minor effect on the model, it follows that the channels in set A are more responsible for causing the model to fail as a whole.
Our experiments on three models trained on known image classification tasks (MNIST-37, CIFAR-10, Imagenette) under the Auto-PGD attack \cite{croce2020ReliableEvaluationAdversarial} revealed that when only certain non-random channels are perturbed by adversarial perturbations, they produce statistically significant changes to the output of the model compared to other channels in the same layer. 
For example, for the model trained on CIFAR-10 the results showed that perturbing only 3 vulnerable channels (out of 16) in the first layer was sufficient to completely break the model ($-98\%$ accuracy). In contrast, the majority of the other random 3-channel combinations had marginal effects on the model’s ability to correctly classify test samples.
Applying the \textit{Adversarial Intervention} framework to the same model but under attacks of a different nature, highlighted the same channels as the most critical to the model despite differences in the magnitude of the effects.

The proposed framework is not just a diagnostic tool for trained models; it also allows for the study of the properties that make certain channels more vulnerable. In fact, after having ranked the channels in order of vulnerability, the natural question that follows is to understand what is making certain channels behave in a certain way. We studied how the vulnerability ranking of channels correlates with the properties of the kernels that represent them. This analysis revealed that the vulnerability ranking of channels has a strong correlation with their corresponding kernel’s $\ell_2$ norm. This work contributes to the understanding of the adversarial perturbations phenomenon by providing a framework
to identify the channels of a layer that can cause a model to be vulnerable as a whole. The experiments revealed that small non-random combinations of channels in the shallow layers can cause significant effects on the model's robustness compared to other random combinations. Further analysis showed that there is a strong
correlation between the magnitude of the values of a kernel and its vulnerability to adversarial perturbations. While this work contributes to understanding the mechanism of adversarial perturbations and does not directly propose a new defense method, but it could lead to the development of model-specific defense strategies in the future.

\section{Related Works}
The vulnerability of neural networks to adversarial perturbations is a known unresolved issue \cite{szegedy2014IntriguingPropertiesNeural, goodfellow2015ExplainingHarnessingAdversarial}. 
Over the years, much of the literature on the topic has focused on the development of adversarial attack algorithms \cite{li2022ReviewAdversarialAttack, long2022SurveyAdversarialAttacks, machado2023AdversarialMachineLearning} and adversarial defense methods \cite{wang2022AdversarialAttacksDefenses,zhao2022AdversarialTrainingMethods,machado2023AdversarialMachineLearning}. 

In parallel, another stream in the literature has been looking for the root causes of such vulnerability.
The phenomenon was first attributed to the highly complex non-linear behavior of deep learning models \cite{szegedy2014IntriguingPropertiesNeural}. However, it was also argued that the commonly used ReLu activations make the networks almost-linear, thus showing that this linear behavior can lead to the observed vulnerability \cite{goodfellow2015ExplainingHarnessingAdversarial}.
The same work also noted the dependence of the vulnerability with the input dimension, which was also confirmed by further theoretical studies \cite{gilmer2018AdversarialSpheres, simon-gabriel2019FirstOrderAdversarialVulnerability} and empirical results \cite{sun2019UnderstandingAdversarialExamples, pan2020IntrinsicDatasetProperties}.

Other works studied the impact of a model's architectural choices to the vulnerability of models \cite{madry2019DeepLearningModels, sun2019UnderstandingAdversarialExamples, huang2021ExploringArchitecturalIngredients, guo2020WhenNASMeets, du2021LearningDiverseStructuredNetworks, mok2021AdvRushSearchingAdversarially}. 
Among these factors, the use of Batch Normalization (BN) was identified as a component that increases a model's performance on ``clean" samples but reduces its robustness to adversarial perturbations \cite{galloway2019BatchNormalizationCause,benz2021BatchNormalizationIncreases,kong2022WhyDoesBatch}.
This work focuses on investigating how the individual feature maps learned by a model during training affect its vulnerability to adversarial perturbations.

Previous works have looked into quantifying the vulnerability of individual neurons \cite{zhang2021InterpretingImprovingAdversarial, bai2022ImprovingAdversarialRobustness, yan2021CIFSImprovingAdversarial, khalooei2022LayerwiseRegularizedAdversarial}.
However, these works mainly focus on sensitivity measures following a perturbation to the input.
In contrast, in this work the objective is to isolate the effect on the model caused by an adversarial perturbation to an individual neuron. 
The intervention-based approach is typical of the causal inference literature and can provide further insights into the mechanisms of a model \cite{moraffah2020CausalInterpretabilityMachine}. 
Measures of sensitivity can only provide information on the effects of a the adversarial perturbations, whereas controlled intervention can help in understanding the causes of such vulnerability. 
Other works have attempted using the Structural Causal Model (SCM) framework \cite{pearl2009CausalityModelsReasoning} to interpret the inner workings of neural networks \cite{narendra2018ExplainingDeepLearning, chattopadhyay2019NeuralNetworkAttributions, janzing2020FeatureRelevanceQuantification}.
The framework proposed in this work is loosely inspired by the SCM framework, however it does not follow its formalism.

\section{Methods}

The \textit{Adversarial Intervention} framework is introduced in this work to understand how adversarial perturbations on individual channels affect a deep learning model (Fig. \ref{fig:adversarial-intervention-framework}).
In short, the framework consists in swapping certain channels in the hidden representation computed for clean samples with the corresponding channels computed for the same samples corrupted by adversarial perturbations. The effect of the perturbation on the selected channel is then evaluated as the difference in the model output between the clean sample and the one modified through \textit{Adversarial Intervention}.
The following sections explain in further detail the operations in this framework (Section \ref{sec:adv-intervention}) and the metrics used to evaluate the effect of the interventions (Section \ref{sec:adv-intervention-metrics}).

\subsection{The \textit{Adversarial Intervention} framework}
\label{sec:adv-intervention}

A Convolutional Neural Network (CNN) is given, formally defined as $g: \mathbb{R}^M\to\mathbb{R}^N$. This network is a made up of a series of sequential operations that process the input $x\in\mathbb{R}^M$ into the output logits vector $z=g(x)\in\mathbb{R}^N$, where $M$ and $N$ indicate the dimensions of the input and output spaces respectively.
Given the logits computed by the model, the class prediction operation is defined as $\hat{y}=\argmax(z)=\argmax(g(x))$.

In this work we are interested in studying the vulnerability to adversarial perturbations of a specific convolutional layer operation in the model. Let us define $L$ the convolutional layer that is the target of our analysis.
Thus, the model $g$ can be split into two sequential functional blocks: 1) $g_A: \mathbb{R}^M \to \mathbb{R}^H$ - the first half of the model, from the input to the target layer $L$ (included); 2) $g_B: \mathbb{R}^H \to \mathbb{R}^N$ - the second half of the model, which takes as input the output of the target layer $L$ and feeds it into the remaining layers of the model until the model output.
In short, the following equivalence holds: $g(x)=g_B(g_A(x))$.

The intermediate representation $h=g_A(x)$ is the output of the target layer and is a tensor of dimension $H=H_{width}\times H_{height} \times H_{channels}$ with $h\in\mathbb{R}^H$.
Now, let us define $x' = x + \eta \in \mathbb{R}^M$ an adversarial example for the input $x$, computed using any arbitrary attack in the literature \cite{li2022ReviewAdversarialAttack, long2022SurveyAdversarialAttacks, machado2023AdversarialMachineLearning}; in the majority of experiments the Auto-PGD attack was employed \cite{croce2020ReliableEvaluationAdversarial}. The tensor $\eta$ is the small ad-hoc noise that is added to $x$ to make it an adversarial example.

For the adversarial sample $x'$ the intermediate representation at the target layer is $h'=g_A(x) \in\mathbb{R}^H$, the model output is $z'=g(x')=g_B(g_A(x'))$, and its class prediction is $\hat{y}'=\argmax(z')$.
The adversarial example is valid if $\hat{y} \neq \hat{y}'$, i.e. if the adversarial perturbations on the input changes the prediction of the model.

Given $h$ (clean intermediate features) and $h'$ (adversarial intermediate features), the \textit{Adversarial Intervention} operation consists in swapping $\gamma$ channels in $h$ with the corresponding channels in $h'$. A graphical representation is in Figure \ref{fig:adversarial-intervention-framework}.
Let us define the following
\begin{itemize}
	\item $H_{channels}$: the number of channels in the intermediate features tensor;
	\item $\gamma$: the number of channels that will be swapped;
	\item $\Omega^{\gamma}$: the set of all possible combinations of channel indices for a given $\gamma$, with size $|\Omega^{\gamma}|=\frac{H_{channels}}{\gamma!(H_{channels}-\gamma)!}$;
	\item $\Phi \in \Omega^{\gamma}$: the set of channel indices that will be swapped by the operation, i.e. $|\Phi|=\gamma$.
\end{itemize}
The intermediate representation $h$ can be written in the form $h=[h_1, h_2, \hdots, h_{H_{channels}}]$ where each element is one of the channels in the tensor, i.e. $h_i \in \mathbb{R}^{H_{width}\times H_{height}} \forall i \in [1, H_{channels}]$. Similarly for the adversarial intermediate features $h'=[h'_1, h'_2, \hdots, h'_{H_{channels}}]$.
For a chosen set of channels $\Phi$, the following mask tensors are defined 
\begin{align}
	m^{\Phi} &= [m_1, m_2, \hdots, m_{H_{channels}}] \\
	\bar{m}^{\Phi} &= 1 - m^{\Phi} \\ 
	\text{with } \quad 
	m^{\Phi}_i &= \begin{cases}
		1 \quad \text{if } i \notin \Phi \\
		0 \quad \text{otherwise}
	\end{cases}
\end{align}
It follows that the \textit{Adversarial Intervention} operation yields the modified tensor of features 
\begin{align}
	h^{\Phi} = h \cdot m^\Phi + h' \cdot \bar{m}^\Phi
\end{align}

The modified tensor of features $h^{\Phi}$ is then inserted as input to the rest of the model (the function $g_B$) to compute the corresponding output. The difference in output and class prediction between the clean sample and the modified sample are used to evaluate the metrics defined in Section \ref{sec:adv-intervention-metrics}.

\subsection{Metrics}
\label{sec:adv-intervention-metrics}

The dataset of clean samples $(x, y) \in \mathcal{X}$ is given, where $x$ is an input image and $y$ is its class label. For every sample $x$ it is possible to compute its adversarial example equivalent $x'$ using any arbitrary attack in the literature. The \textit{Adversarial Intervention} for a given set of channels $\Phi$ yields the intermediate feature vector $h^\Phi$ as defined in the previous section.
It is assumed that the samples in $\mathcal{X}$ are all correctly classified by the given model (i.e. $\hat{y}=y, \, \forall (x, y) \in \mathcal{X}$), in order to reduce the confounding effect in the analysis that may come from the model's inability of classifying certain samples even when they are not perturbed.
For a given value of $\gamma$ (number of channels to modify), let us define the subset $\Omega^\gamma_j$ of all the combinations of $\gamma$ channels that include a specific channel $j$; formally
\begin{equation}
	\Omega^\gamma_j = \left\{ \Phi : j \in \Phi \right\} \subset \Omega^\gamma
\end{equation}

It is now possible to quantify the effect that adversarial perturbations on individual channel(s) have on the output of a model. Specifically, the focus is on the effect on the output logits vector and the prediction of the model. The following metrics are defined.

\begin{definition}[$AEL_{\Phi}$: Average Effect on Logits of $\Phi$]
	Given a data distribution $\mathcal{X}$, the Average Effect on Logits (AEL) of the \textit{Adversarial Intervention} on the channels $\Phi$ is the expected value of the $\ell_2$-norm of the difference between the output of the model for a clean sample and a sample modified through \textit{Adversarial Intervention}, i.e.
	\begin{align}
		AEL_{\Phi} = \mathbb{E}_{\mathcal{X}} \left[|g_B(h) - g_B(h^\Phi)|_2 \right]
	\end{align}
\end{definition}

The metric $AEL_{\Phi}$ quantifies the average effect on the logits when the specific set of channel indices in $\Phi$ are adversarially perturbed, while the rest are left unpertrubed.

\begin{definition}[$AEL_{j}^{\gamma}$: Average Effect on Logits of channel $j$]
	For a fixed value of $\gamma$, the Average Effect on Logits (AEL) of applying the \textit{Adversarial Intervention} on the $j$-th channel is the mean value of $AEL_\Phi$ over all the sets of $\Omega^\gamma$ that include $j$, i.e.
	\begin{align}
		AEL_j^\gamma &= \frac{1}{|\Omega_j^\gamma|} \sum_{\Phi \in \Omega^\gamma_j} AEL_{\Phi} 
	\end{align}
\end{definition}

The metric $AEL_{j}^{\gamma}$ quantifies the average effect on the logits when channel $j$ and other $\gamma - 1$ channels are adversarially perturbed, while the rest are left unperturbed.
To define the effect on the accuracy of the perturbations, we first introduce a correctness metric function 

\begin{equation}
	\mathds{1}(a,b) = 
	\begin{cases}
		1 &\quad \text{if } a = b \\
		0 &\quad \text{otherwise}
	\end{cases}
\end{equation}

which allows to define the average accuracy of a model as the expectation on the dataset. 
Recall that $y$ is the true label of the input, while $\hat{y}$ is the prediction of the model to a clean sample. Hence, we define $\hat{y}^{\Phi}$ as the prediction of the model when the channels in $\Phi$ have been swapped.
By assumption, all the samples in the dataset $\mathcal{X}$ are correctly classified by the model, hence its average accuracy on the dataset is $\mathbb{E}_{\mathcal{X}}\left[\mathds{1}(y, \hat{y})\right]=1$.

\begin{definition}[$AEA_{\Phi}$: Average Effect on Accuracy of $\Phi$]
	Given a data distribution $\mathcal{X}$ of clean samples correctly classified by the model, the Average Effect on Accuracy (AEA) of the \textit{Adversarial Intervention} on the channels $\Phi$ is the expected value of the change in the model accuracy on the dataset, i.e.
	\begin{align}
		AEA_{\Phi} 	&= \mathbb{E}_{\mathcal{X}}\left[\mathds{1}(y, \hat{y})\right] - \mathbb{E}_{\mathcal{X}} \left[ \mathds{1}(y, \hat{y}^\Phi) \right] \\
		&= 1 - \mathbb{E}_{\mathcal{X}} \left[ \mathds{1}(y, \hat{y}^\Phi) \right]
	\end{align}
\end{definition}

The metric $AEA_{\Phi}$ quantifies the average effect on the average accuracy of the model when the $\gamma$ channels in $\Phi$ are adversarially perturbed, while the rest are left unperturbed.

\begin{definition}[$AEA_{j}^{\gamma}$: Average Effect on Accuracy of channel $j$]
	For a fixed value of $\gamma$, the Average Effect on Accuracy (AEA) of applying the \textit{Adversarial Intervention} on the $j$-th channel is the mean value of $AEA_\Phi$ over all the sets of $\Omega^\gamma$ that include $j$, i.e.
	\begin{align}
		AEA_j^\gamma &= \frac{1}{|\Omega_j^\gamma|} \sum_{\Phi \in \Omega^\gamma_j} AEA_{\Phi} 
	\end{align}
\end{definition}

The metric $AEA_{j}^{\gamma}$ quantifies the average effect on the average accuracy of the model when the $j$-th channel of the target layer and other $\gamma - 1$ channels are adversarially perturbed, while the rest are left unperturbed.

\section{Results}
\label{sec:results}

\subsection{Experimental Setup}
\label{sec:experimental-setup}

\begin{table}[t]
	\centering
	\begin{tabular}{l|l|l|l}
		& \textit{MNIST-37} & \textit{CIFAR-10}  & \textit{Imagenette} \\ \hline
		model & 4-layer CNN	 & ResNet20  &EfficientNetV2\\ \hline
		examined layer & 1st conv. layer & 1st conv. layer  &1st conv. layer\\ \hline
		output channels of examined layer & 16 & 16  &32\\ \hline
		accuracy (clean) & 98.92\% & 92.27\%  & 96.07\% \\ \hline
		accuracy (APGD) & 0.00\% & 0.00\% & 0\% \\ \hline
		accuracy (C\&W) & - &  0.37\% & - \\ \hline
		accuracy (HSJ) & - &  1.86\% & - 
	\end{tabular}
	\caption{Summary of models used in the experiments. Accuracy clean is computed on test set. Accuracy after attack is computed on subset of correctly classified samples. Attacks: Auto-PGD (APGD); Carlini \& Wagner L2 (C\&W); Hop-Skip Jump (HSJ).}
	\label{tab:models}
\end{table}

Experiments have been carried out on three datasets: MNIST-37, CIFAR-10 \cite{cifar10}, and Imagenette \cite{imagenette}.
MNIST-37 is a subset of the classic MNIST \cite{deng2012mnist} that includes only the digits ``3" and ``7". 
For MNIST-37 we use a simple CNN with 4 convolutional layers and 1 fully connected layer; the target layer is the first layer (16 channels). This dataset is relatively simple and acted as a proof of concept for the analysis.
For CIFAR-10 we use the ResNet-20 model \cite{he2016DeepResidualLearning}; the target layer is the first layer (16 channels). This dataset has a moderate complexity and allowed us to explore variations on the type of attack used.
For Imagenette we use the EfficientNetV2 model \cite{tan2021EfficientNetV2SmallerModels}; the target layer is the first convolutional layer (32 channels). This dataset has a higher complexity and also requires more computational time during the analysis.
These datasets and model architectures have been chosen in order to be able to train models with very high accuracy on unseen samples, which should focus the attention of our analysis on the vulnerability to adversarial perturbations rather than mistakes in generalization. 
Following this principle, the analysis with the \textit{Adversarial Intervention} framework has been carried out on samples that are unknown to the model (i.e. not been used during training) and that the models correctly classify.

Adversarial examples for the samples in the datasets have been computed using the Auto-PGD (APGD) \cite{croce2020ReliableEvaluationAdversarial} with a maximum perturbation of the $\ell_\infty$ norm of $8/255$ for CIFAR-10 and Imagenette, and 0.30 for MNIST-37.
Auto-PGD is a strong state-of-the-art adversarial attack that reduces the effect that hyperparameter choices have on the attack.
Further experiments have been carried out on the model trained on CIFAR-10 using additional attacks, such as the Carlini \& Wagner L2 attack (C\&W) \cite{carlini2017EvaluatingRobustnessNeural} and   the Hop-Skip Jump (HSJ) attack \cite{chen2020HopSkipJumpAttackQueryEfficientDecisionBased}. These two attacks are of a different nature than APGD and follow a different logic, thus allowing to inspect how different attacks impact the internal representations of the same model.
For all the setups, many values of $\gamma$  have been considered. For MNIST-37 and CIFAR-10, all the possible combinations were examined ($\gamma \in [1,16]$), whereas for the Imagenette experiments we only considered $\gamma \in [1,4]$ due to the exponentially growing number of channel combinations and computational limitations. Nonetheless, as the experiments on the smaller showed, the results obtained for smaller values of $\gamma$ are a good indication of the results for combinations of more channels. Moreover, the more channels are perturbed at the same time, the more the effects of an individual channel become confounded.

The experiments were programmed using Python, the Tensorflow 2 framework for deep learning \cite{abadi2016TensorflowSystemLargescale}, and the Adversarial Robustness Toolbox (ART) implementation of the adversarial attacks \cite{art2018}. To study the behavior of naturally trained models, no defense methods have been implemented during training (e.g. adversarial training) nor inference.
Further details on the models are summarized in Table \ref{tab:models}.

\subsection{Analyzing the effect of feature-level adversarial perturbations}

The experiments indicate that for the three models there exist certain channels in the first layer that can cause severe disruptions the models when perturbed by adversarial perturbations.
Starting with applying \textit{Adversarial Intervention} to a single channel (i.e. $\gamma=1$), it is possible to see that there exist certain channels that have significantly higher average effects on the output of the model (Figure \ref{fig:gamma1} in Appendix \ref{apx:plots}).
For a simple binary classification problem like MNIST-37, perturbing a single channel does not have major effects on the model's ability to correctly classify a sample.
On the contrary, for a slightly more difficult problem such as CIFAR-10, the results show that perturbations to certain individual channels can cause the model accuracy to drop by up to 19\%.

As expected, as we increase the number of channels that are modified at the same time, the effect of the perturbations on the models increases.
For the simple MNIST-37 case, serious effects on the model's classification ability emerge when almost half of the channels in the first layer are perturbed through \textit{Adversarial Intervention} (Figure \ref{apx:fig:mnist:gamma7}).
The results on the CIFAR-10 appear more interesting, as there are certain channels that stand out way more for their effects on the model. 
Figure \ref{fig:cifar:gamma3} reports the results obtained when perturbing combinations of 3 channels in the first layer of the model.
In this scenario, perturbing 3 specific channels is sufficient to completely break the model, with drops of average accuracy of more than $90\%$ in certain cases.
The difference in the effects caused by the top and bottom $\Phi$ combinations is truly significant (Figure \ref{fig:cifar:gamma3}a). This means that there exist certain channels that are way less robust to the adversarial perturbations, while others barely affect the model as a whole when perturbed.
The histogram collecting the $AEA_\Phi$ of all the combinations for $\gamma=3$ further reveals that the majority of combinations have relatively small effects on the model, whereas a few can severely affect the model (Figure \ref{apx:fig:cifar:gamma3}c).
The metrics with channel-wise stratification displayed in Figure \ref{fig:cifar:gamma3}b reveal the significant difference in the effects caused by the individual channels. 
It is noteworthy that the channels with the highest individual effects for $\gamma=3$ are the same that had already been identified with the single-channel perturbation ($\gamma=1$).

Similar observations can be made for the experiments on the Imagenette dataset, which is a more complex classification task due to the nature of the data. The model used in these experiments (EfficientNetV2) has more trainable parameters and the target layer has 32 channels, double the number of the previous cases. Given the higher number of channels, a greater redundancy is expected in the features learned by the model. This means that more channels would have to be corrupted to have significant effects on the model. Nonetheless, the results in Figure \ref{fig:imagenette:gamma4:top10} show that certain combinations of 4 channels are sufficient to have very significant effects on the output logits and to reduce the accuracy of the model by more than $-50\%$. The single channel effects in Figure \ref{fig:imagentte:gamma4:sc} further reveal the significant differences in the effects caused by individual channels when they are perturbed by adversarial perturbations.

As the value of $\gamma$ increases, it becomes harder to highlight the effect of a specific channel due to the confounding effect of channel interactions. Nonetheless, looking at the ranking of individual channels over the examined values of $\gamma$ reveals that the ranking of a given channel is relatively stable even when other channels are perturbed (Figure \ref{fig:channel-ranking}). 
Additional plots for the experiments can be found in the Appendix \ref{apx:plots}.

\newcommand{\resultsPlotSize}{.48}
\begin{figure}
	\centering
	\begin{subfigure}{\resultsPlotSize\linewidth}
		\includegraphics[width=\linewidth]{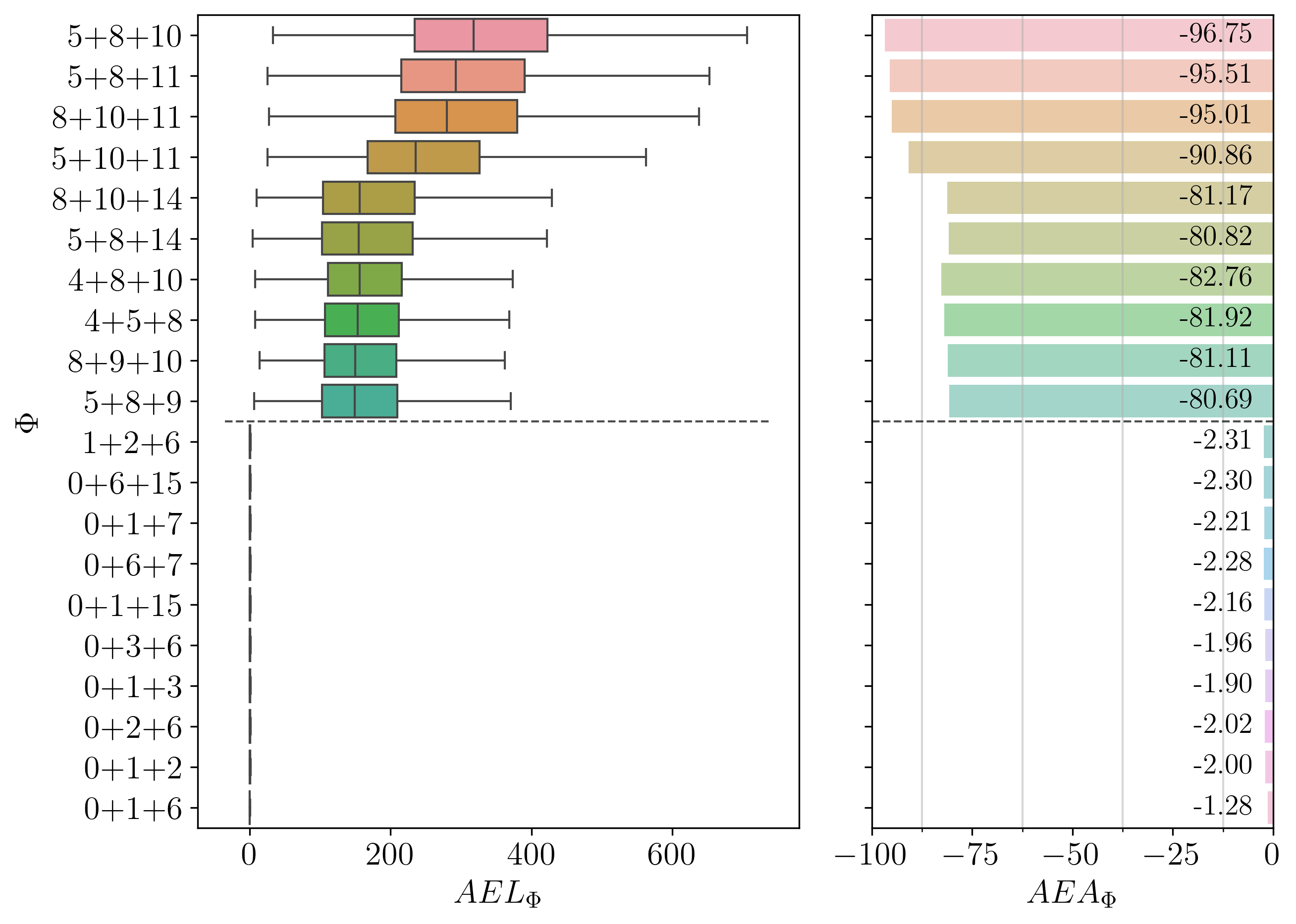}
		\caption{}
		\label{fig:cifar:gamma3:top10}
	\end{subfigure}
	\quad
	\begin{subfigure}{\resultsPlotSize\linewidth}
		\includegraphics[width=\linewidth]{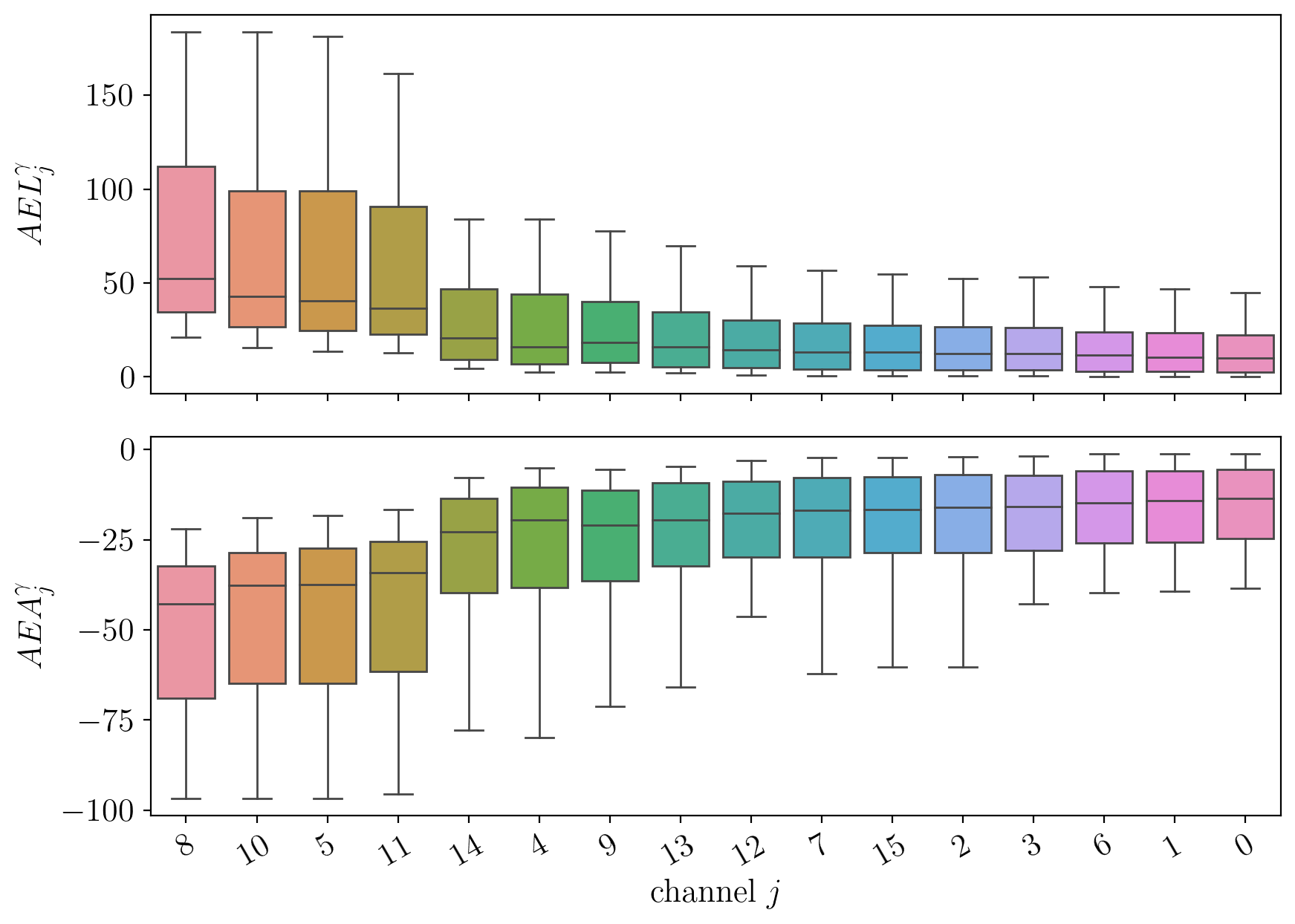}
		\caption{}
		\label{fig:cifar:gamma3:sc}
	\end{subfigure}
	\caption{
		Results of \textit{Adversarial Intervention} (Auto-PGD) on CIFAR-10 model for $\gamma=3$. (a) Top and bottom 10 channel combinations, ranked by $AEL_\Phi$. Certain combinations $\Phi$ can disrupt the model performance almost completely, while others barely affect it. (b) Channel-wise effects on logits and accuracy. The effect of certain channels is significantly different than others.
	}
	\label{fig:cifar:gamma3}
\end{figure}

\begin{figure}
	\centering
	\begin{subfigure}{\resultsPlotSize\linewidth}
		\includegraphics[width=\linewidth]{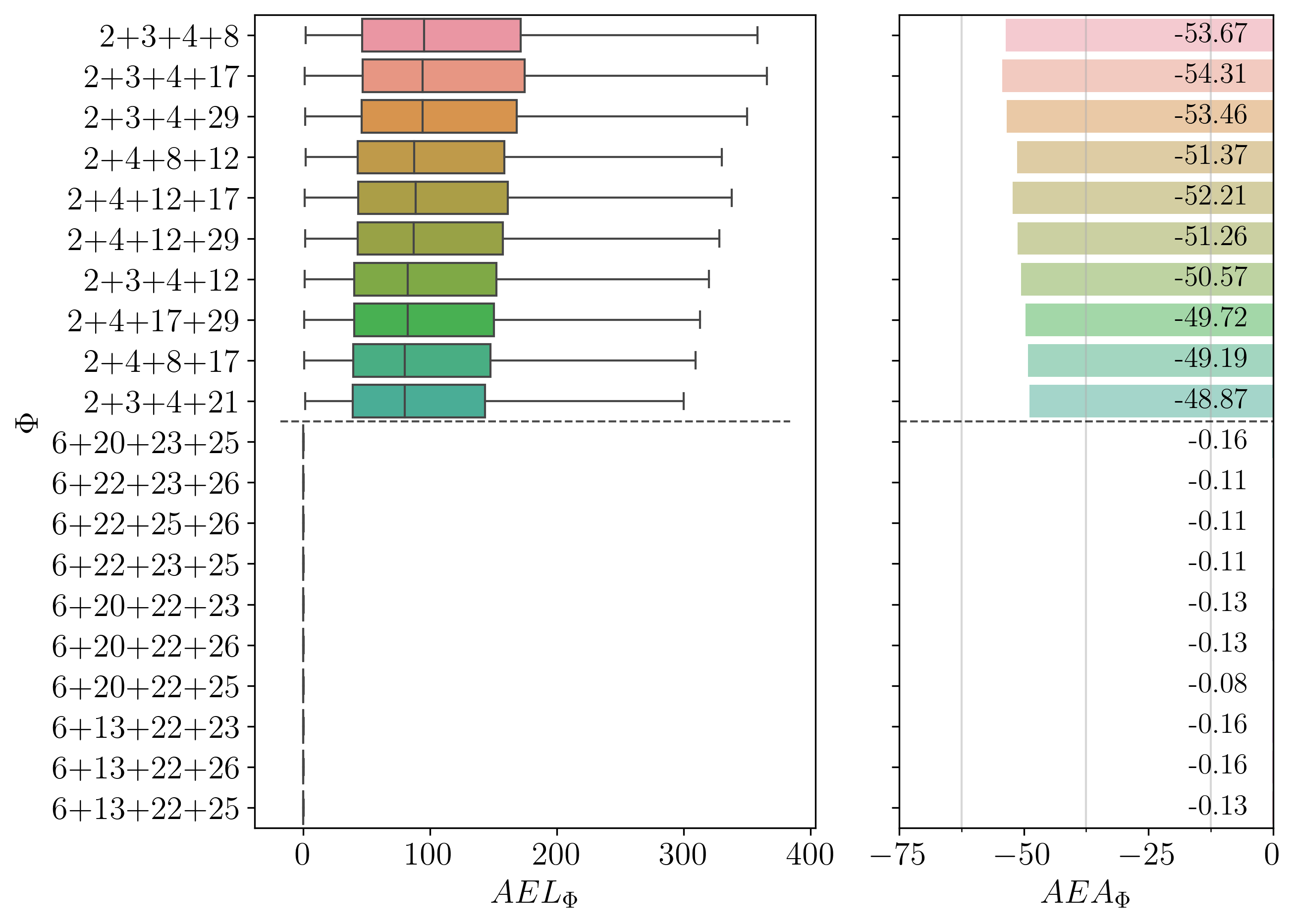}
		\caption{}
		\label{fig:imagenette:gamma4:top10}
	\end{subfigure}
	\quad
	\begin{subfigure}{\resultsPlotSize\linewidth}
		\includegraphics[width=\linewidth]{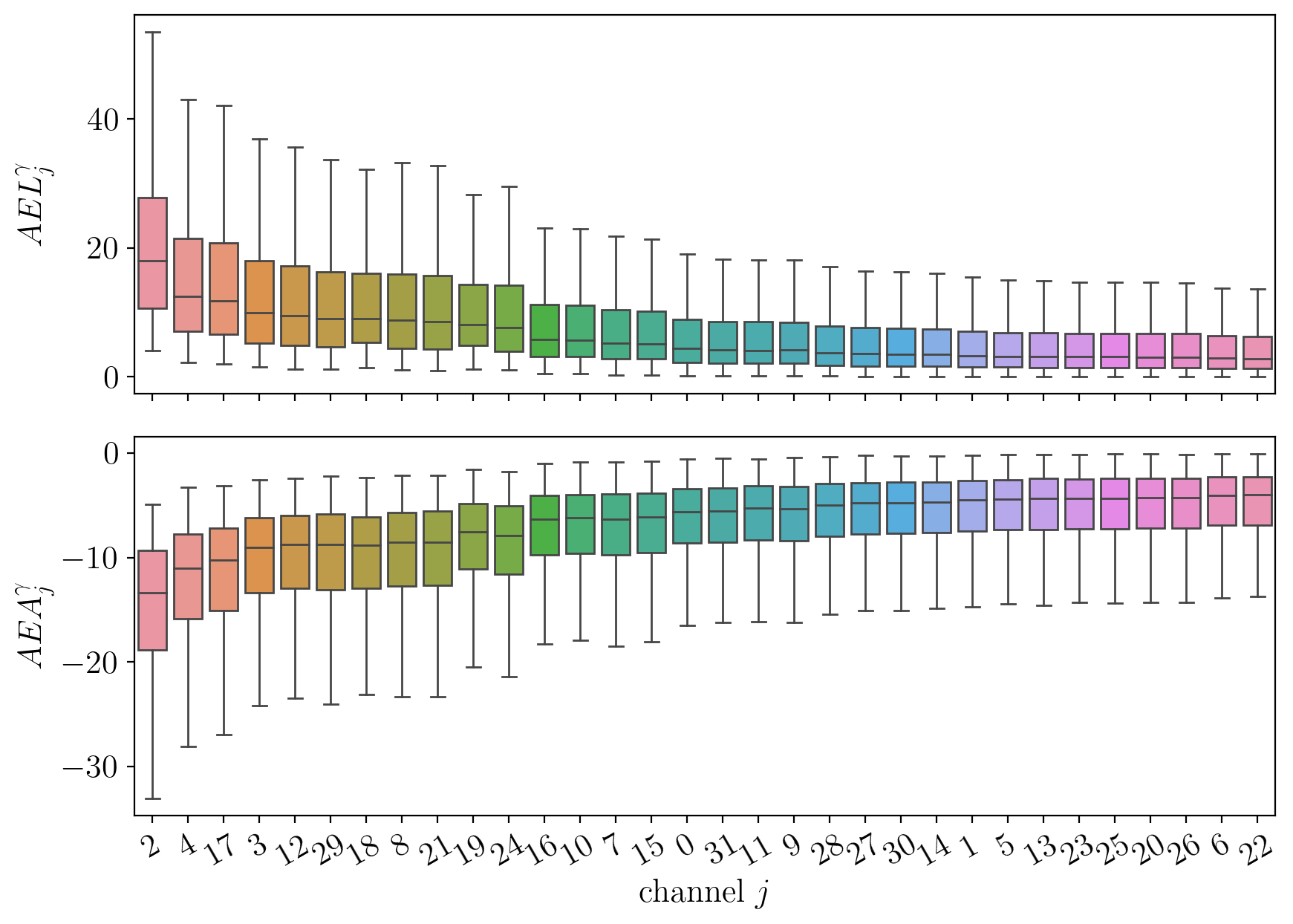}
		\caption{}
		\label{fig:imagentte:gamma4:sc}
	\end{subfigure}
	\caption{
		Results of \textit{Adversarial Intervention} (Auto-PGD) on Imagenette for $\gamma=4$. (a) Top and bottom 10 channel combinations, ranked by $AEL_\Phi$. (b) Channel-wise effects on logits and accuracy. 
		The plots show a clear distinction in the model's response to top- and bottom-ranking combinations, and highlight specific channels that significantly disrupt the model's output.
	}
	\label{fig:cifar:gamma4}
\end{figure}

\begin{figure}
	\centering
	\includegraphics[width=\linewidth]{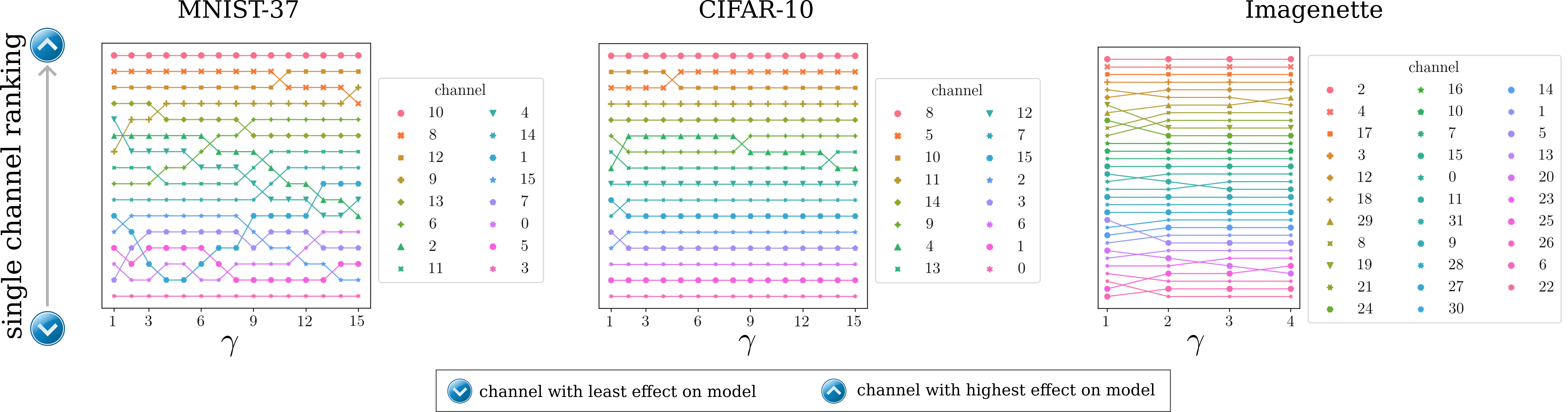}
	\caption{Channel ranking over values of $\gamma$ for \textit{Adversarial Intervention} with Auto-PGD. In all three models, the ranking of a channel in terms of its average effect on the model output is relatively stable as the value of $\gamma$ increases.}
	\label{fig:channel-ranking}
\end{figure}

\subsection{Inspecting the effect of attacks with different nature}

\begin{figure}[t]
	\centering
	\includegraphics[width=\linewidth]{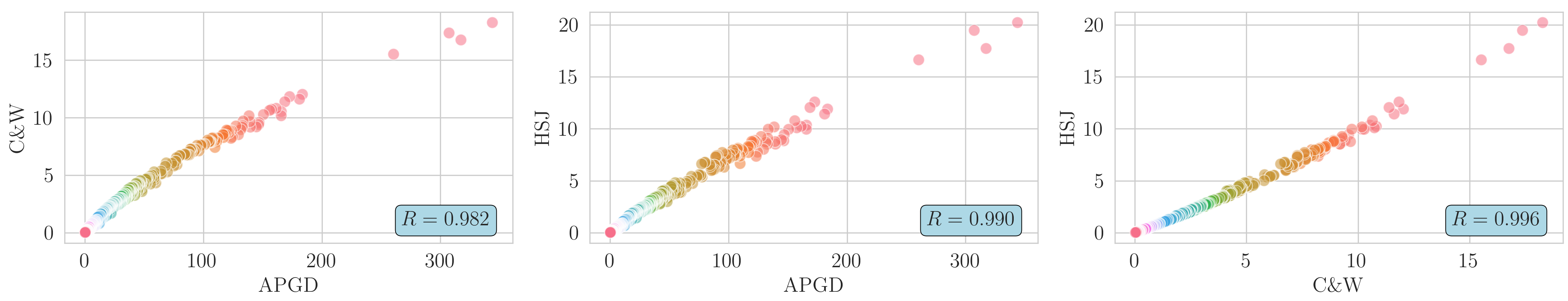}
	\caption{Comparison among the effects on logits of individual channel combinations for attacks of different nature on the same ResNet20 model trained on CIFAR-10. The values in the axes indicate the Average Effect on Logits (AEL) when a given channel combination is perturbed through \textit{Adversarial Intervention} with the attack indicated on the axis label. Each point corresponds to channel combination. The results in these plots are for combinations of $\gamma=3$ channels.}
	\label{fig:cifar:cmpr-atks:g3}
\end{figure}

To inspect whether the channel combinations that are most responsible for the vulnerability of the model depend on the type of attack used, we repeated the attacks on the ResNet20 trained on CIFAR-10 with two other attacks of a different nature: Carlini \& Wagner attack (C\&W) \cite{carlini2017EvaluatingRobustnessNeural} and Hop-Skip Jump attack (HSJ) \cite{chen2020HopSkipJumpAttackQueryEfficientDecisionBased}. 
Auto-PGD and C\&W are both white box attacks but their algorithms follow different procedures to generate the adversarial examples. In contrast, the HSJ attack is a black-box attack that only uses the input and output information to compute the samples.
Firstly it was noted how the two new attacks require a higher number of channels ($\gamma$) to be perturbed to significantly disrupt the accuracy of the model ($AEA$). For example, for Auto-PGD certain combinations of only $\gamma=3$ channels were able to already reduce the accuracy by more than $90\%$. On the other hand, for C\&W and HSJ such intense disruptions of the accuracy were only observed for very high values of $\gamma$, which perturbed almost all the channels available (Figures \ref{apx:fig:cifar:gamma3:cwl2}-\ref{apx:fig:cifar:gamma14:hsj}). This shows how different algorithms lead to perturbations that affect the inner representations in very different ways. 
Despite the magnitude of the effects on the accuracy being low, looking into the average effect on the logits ($AEL$) reveals how there are still certain channels that cause statistically significantly higher disruptions than others. Interestingly, the ranking of channel combinations based on their effect on the logits is mostly unvaried among the three attacks.
We compared the $AEL_\Phi$ for one attack with the $AEL_\Phi$ obtained with another attack (Figure \ref{fig:cifar:cmpr-atks:g3}) and found that, despite the different magnitudes, there is a very high correlation between the values of the effects. In short, the channel combinations that are deemed most vulnerable for one attack are the same when inspecting the other attacks too.

\subsection{What makes a channel more vulnerable than others?}

The data collected for these models clearly suggests that certain channels in the first layers can cause tremendous disruptions to a model if they are perturbed by adversarial perturbations, even when all the features extracted by the other channels in the same layer are ``correct" (i.e. unperturbed).
Following this realization, the next step is to understand what is making those channels more prone to cause such effects on the models.

Each channel of a convolutional layer is associated with a kernel of trainable weights that have been learned by the model during the training phase. The kernels process the input independently from one another and each one corresponds to a certain feature that the model has learned to achieve its task.
If a channel is more vulnerable to adversarial perturbations, one possible explanation is that the model has learned to recognize certain features that are useful to achieve the learning objective but are not robust nor meaningful.

\begin{figure}
	\centering
	\includegraphics[width=1.\linewidth]{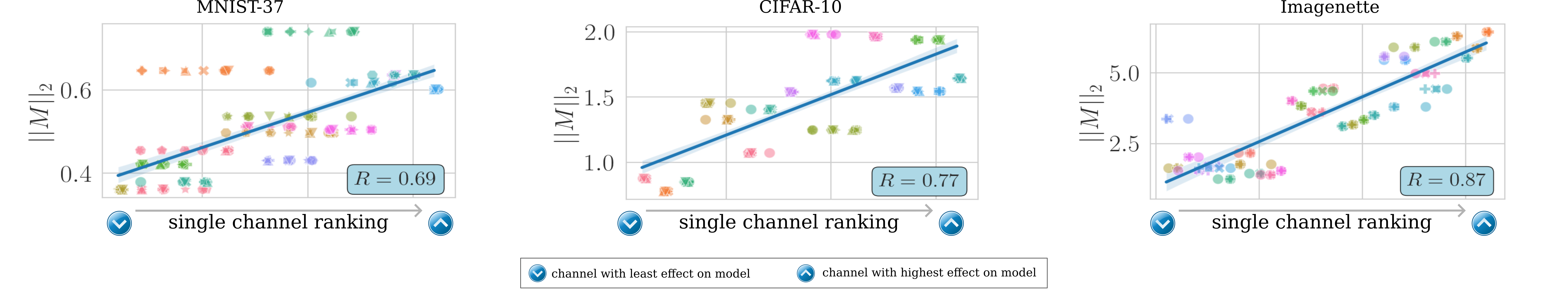}
	\caption{Correlation between vulnerability rank of a channel and properties of its associated kernel $M$: $\ell_2$ norm of the kernel and standard deviation of values. Each point in the plots represents the ranking of a channel for a different value of $\gamma\in [1, 15]$ (color=channel; shape=$\gamma$). In text box: Pearson's correlation coefficient $R$ associated with linear least-squares regression.}
	\label{fig:rank-correlation}
\end{figure}

We explored how certain descriptive statistics of the kernels correlate with the vulnerability ranking of that channel.
The values of the $\ell_2$ norm of a kernel revealed positive correlations with the channel's ranking (Figure \ref{fig:rank-correlation}), appearing to be a potential cause for its vulnerability.
Intuitively, this finding makes sense as a bigger kernel implies that the input values will be multiplied by numerically bigger coefficients. In turn, this means that if there is some adversarial noise in the input, this will also be enlarged by the multiplication.
Further experiments on more datasets and models will be necessary to reinforce the findings on what makes a channel more vulnerable than others.

\section{Conclusions}
\label{sec:conclusions}

This study addresses the vulnerability of Convolutional Neural Network (CNN) models to adversarial perturbations, with a particular focus on understanding the role of feature representations learned by the model. Previous work has highlighted the importance of shallow layers in the model's vulnerability, and this research builds upon that by examining individual channels within these layers.
While existing studies have quantified channel vulnerability in terms of their sensitivity to adversarial perturbations, they primarily focus on the effects of perturbations rather than investigating the underlying causes. In contrast, this study investigates the causes of channel vulnerability and proposes a framework called \textit{Adversarial Intervention} to shed light on this phenomenon.

The \textit{Adversarial Intervention} framework is inspired by causal inference literature and involves modifying clean hidden representations by substituting certain channels with their perturbed counterparts. By analyzing the resulting changes in the model's output, the vulnerability of individual channels can be revealed. Specifically, if perturbing a small set of channels has a significant disruptive effect on the model's behavior, while perturbing another equally small set has minimal impact, it suggests that the channels in the former set are more vulnerable and contribute to the overall vulnerability of the model.
Experiments conducted on models trained on MNIST-37, CIFAR-10, and Imagenette datasets confirm that certain channels exhibit statistically significant higher effects on the model's output compared to others. For instance, perturbing only three vulnerable channels in the first layer of the CIFAR-10 model resulted in a drastic drop in accuracy ($-98\%$). In contrast, most random combinations of three channels had minimal effects on the model's classification performance.
Applying the framework on the same model trained on the CIFAR-10 dataset but with attacks of a different nature revealed that different attacks affect the models with distinct mechanisms, but share the channels which cause the highest disruption to the model's output.
Importantly, the proposed framework not only serves as a diagnostic tool for trained models but also enables the study of properties that contribute to channel vulnerability. In this regard, the study examines the correlation between vulnerability ranking and kernel properties. The analysis reveals a strong correlation between the vulnerability ranking of channels and the $\ell_2$ norm of their corresponding kernels.

Overall, this research contributes to the understanding of the adversarial perturbations phenomenon by introducing a framework to identify vulnerable channels within a layer, which can ultimately render the entire model vulnerable. The findings demonstrate that non-random combinations of channels in shallow layers can have substantial effects on model robustness. Additionally, the correlation between vulnerability and kernel properties suggests the potential for developing targeted defense methods to mitigate model vulnerability.
While this study does not propose new defense methods, it lays the groundwork for the development of more efficient ad-hoc defense strategies that address individual channels to rectify model vulnerability. Further research is needed to explore and validate these findings on larger-scale experiments and different datasets, expanding our understanding of adversarial perturbations and enabling the development of more robust defense mechanisms.

\section*{Acknowledgments}
This study was partially funded by the Biomedical Research Council of the Agency for Science, Technology and Research (A*STAR), Singapore.

\bibliographystyle{unsrt}  
\bibliography{references}  

\begin{thebibliography}{10}

\bibitem{qiu2019ReviewArtificialIntelligence}
Shilin Qiu, Qihe Liu, Shijie Zhou, and Chunjiang Wu.
\newblock Review of {{Artificial Intelligence Adversarial Attack}} and
  {{Defense Technologies}}.
\newblock {\em Applied Sciences}, 9(5):909, January 2019.

\bibitem{wang2022AdversarialAttacksDefenses}
Jia Wang, Chengyu Wang, Qiuzhen Lin, Chengwen Luo, Chao Wu, and Jianqiang Li.
\newblock Adversarial attacks and defenses in deep learning for image
  recognition: {{A}} survey.
\newblock {\em Neurocomputing}, 514:162--181, December 2022.

\bibitem{szegedy2014IntriguingPropertiesNeural}
Christian Szegedy, Wojciech Zaremba, Ilya Sutskever, Joan Bruna, Dumitru Erhan,
  Ian Goodfellow, and Rob Fergus.
\newblock Intriguing properties of neural networks.
\newblock {\em arXiv:1312.6199 [cs]}, February 2014.

\bibitem{goodfellow2015ExplainingHarnessingAdversarial}
Ian~J. Goodfellow, Jonathon Shlens, and Christian Szegedy.
\newblock Explaining and {{Harnessing Adversarial Examples}}.
\newblock {\em arXiv:1412.6572 [cs, stat]}, March 2015.

\bibitem{gilmer2019AdversarialExamplesAre}
Justin Gilmer, Nicolas Ford, Nicholas Carlini, and Ekin Cubuk.
\newblock Adversarial {{Examples Are}} a {{Natural Consequence}} of {{Test
  Error}} in {{Noise}}.
\newblock In {\em Proceedings of the 36th {{International Conference}} on
  {{Machine Learning}}}, pages 2280--2289. {PMLR}, May 2019.

\bibitem{ilyas2019AdversarialExamplesAre}
Andrew Ilyas, Shibani Santurkar, Dimitris Tsipras, Logan Engstrom, Brandon
  Tran, and Aleksander Madry.
\newblock Adversarial {{Examples Are Not Bugs}}, {{They Are Features}}.
\newblock In {\em Advances in {{Neural Information Processing Systems}}},
  volume~32. {Curran Associates, Inc.}, 2019.

\bibitem{paknezhad2022ExplainingAdversarialVulnerability}
Mahsa Paknezhad, Cuong~Phuc Ngo, Amadeus~Aristo Winarto, Alistair Cheong,
  Chuen~Yang Beh, Jiayang Wu, and Hwee~Kuan Lee.
\newblock Explaining adversarial vulnerability with a data sparsity hypothesis.
\newblock {\em Neurocomputing}, January 2022.

\bibitem{zhang2021AdversarialPerturbationDefense}
Xingwei Zhang, Xiaolong Zheng, and Wenji Mao.
\newblock Adversarial {{Perturbation Defense}} on {{Deep Neural Networks}}.
\newblock {\em ACM Computing Surveys}, 54(8):159:1--159:36, October 2021.

\bibitem{bai2022ImprovingAdversarialRobustness}
Yang Bai, Yuyuan Zeng, Yong Jiang, Shu-Tao Xia, Xingjun Ma, and Yisen Wang.
\newblock Improving {{Adversarial Robustness}} via {{Channel-wise Activation
  Suppressing}}.
\newblock In {\em International {{Conference}} on {{Learning
  Representations}}}, February 2022.

\bibitem{yan2021CIFSImprovingAdversarial}
Hanshu Yan, Jingfeng Zhang, Gang Niu, Jiashi Feng, Vincent Tan, and Masashi
  Sugiyama.
\newblock {{CIFS}}: {{Improving Adversarial Robustness}} of {{CNNs}} via
  {{Channel-wise Importance-based Feature Selection}}.
\newblock In {\em Proceedings of the 38th {{International Conference}} on
  {{Machine Learning}}}, pages 11693--11703. {PMLR}, July 2021.

\bibitem{khalooei2022LayerwiseRegularizedAdversarial}
Mohammad Khalooei, Mohammad~Mehdi Homayounpour, and Maryam Amirmazlaghani.
\newblock Layer-wise {{Regularized Adversarial Training}} using {{Layers
  Sustainability Analysis}} ({{LSA}}) framework, February 2022.

\bibitem{pearl2009CausalityModelsReasoning}
Judea Pearl.
\newblock {\em Causality Models, Reasoning, and Inference}.
\newblock {Cambridge University Press}, {Cambridge}, 2009.

\bibitem{chattopadhyay2019NeuralNetworkAttributions}
Aditya Chattopadhyay, Piyushi Manupriya, Anirban Sarkar, and Vineeth~N.
  Balasubramanian.
\newblock Neural {{Network Attributions}}: {{A Causal Perspective}}.
\newblock {\em arXiv:1902.02302 [cs, stat]}, July 2019.

\bibitem{croce2020ReliableEvaluationAdversarial}
Francesco Croce and Matthias Hein.
\newblock Reliable evaluation of adversarial robustness with an ensemble of
  diverse parameter-free attacks.
\newblock {\em arXiv:2003.01690 [cs, stat]}, August 2020.

\bibitem{li2022ReviewAdversarialAttack}
Yao Li, Minhao Cheng, Cho-Jui Hsieh, and Thomas C.~M. Lee.
\newblock A {{Review}} of {{Adversarial Attack}} and {{Defense}} for
  {{Classification Methods}}.
\newblock {\em The American Statistician}, January 2022.

\bibitem{long2022SurveyAdversarialAttacks}
Teng Long, Qi~Gao, Lili Xu, and Zhangbing Zhou.
\newblock A survey on adversarial attacks in computer vision: {{Taxonomy}},
  visualization and future directions.
\newblock {\em Computers \& Security}, 121:102847, October 2022.

\bibitem{machado2023AdversarialMachineLearning}
Gabriel~Resende Machado, Eug{\^e}nio Silva, and Ronaldo~Ribeiro Goldschmidt.
\newblock Adversarial {{Machine Learning}} in {{Image Classification}}: {{A
  Survey Towards}} the {{Defender}}'s {{Perspective}}.
\newblock {\em ACM Computing Surveys}, 55(1):1--38, January 2023.

\bibitem{zhao2022AdversarialTrainingMethods}
Weimin Zhao, Sanaa Alwidian, and Qusay~H. Mahmoud.
\newblock Adversarial {{Training Methods}} for {{Deep Learning}}: {{A
  Systematic Review}}.
\newblock {\em Algorithms}, 15(8):283, August 2022.

\bibitem{gilmer2018AdversarialSpheres}
Justin Gilmer, Luke Metz, Fartash Faghri, Samuel~S. Schoenholz, Maithra Raghu,
  Martin Wattenberg, and Ian Goodfellow.
\newblock Adversarial {{Spheres}}, September 2018.

\bibitem{simon-gabriel2019FirstOrderAdversarialVulnerability}
Carl-Johann {Simon-Gabriel}, Yann Ollivier, Leon Bottou, Bernhard
  Sch{\"o}lkopf, and David {Lopez-Paz}.
\newblock First-{{Order Adversarial Vulnerability}} of {{Neural Networks}} and
  {{Input Dimension}}.
\newblock In {\em Proceedings of the 36th {{International Conference}} on
  {{Machine Learning}}}, pages 5809--5817. {PMLR}, May 2019.

\bibitem{sun2019UnderstandingAdversarialExamples}
Ke~Sun, Zhanxing Zhu, and Zhouchen Lin.
\newblock Towards {{Understanding Adversarial Examples Systematically}}:
  {{Exploring Data Size}}, {{Task}} and {{Model Factors}}, February 2019.

\bibitem{pan2020IntrinsicDatasetProperties}
Jeffrey~Z. Pan and Nicholas Zufelt.
\newblock On {{Intrinsic Dataset Properties}} for {{Adversarial Machine
  Learning}}, May 2020.

\bibitem{madry2019DeepLearningModels}
Aleksander Madry, Aleksandar Makelov, Ludwig Schmidt, Dimitris Tsipras, and
  Adrian Vladu.
\newblock Towards {{Deep Learning Models Resistant}} to {{Adversarial
  Attacks}}.
\newblock {\em arXiv:1706.06083 [cs, stat]}, September 2019.

\bibitem{huang2021ExploringArchitecturalIngredients}
Hanxun Huang, Yisen Wang, Sarah Erfani, Quanquan Gu, James Bailey, and Xingjun
  Ma.
\newblock Exploring {{Architectural Ingredients}} of {{Adversarially Robust
  Deep Neural Networks}}.
\newblock In {\em Advances in {{Neural Information Processing Systems}}},
  volume~34, pages 5545--5559. {Curran Associates, Inc.}, 2021.

\bibitem{guo2020WhenNASMeets}
Minghao Guo, Yuzhe Yang, Rui Xu, Ziwei Liu, and Dahua Lin.
\newblock When {{NAS Meets Robustness}}: {{In Search}} of {{Robust
  Architectures Against Adversarial Attacks}}.
\newblock In {\em Proceedings of the {{IEEE}}/{{CVF Conference}} on {{Computer
  Vision}} and {{Pattern Recognition}}}, pages 631--640, 2020.

\bibitem{du2021LearningDiverseStructuredNetworks}
Xuefeng Du, Jingfeng Zhang, Bo~Han, Tongliang Liu, Yu~Rong, Gang Niu, Junzhou
  Huang, and Masashi Sugiyama.
\newblock Learning {{Diverse-Structured Networks}} for {{Adversarial
  Robustness}}.
\newblock In {\em Proceedings of the 38th {{International Conference}} on
  {{Machine Learning}}}, pages 2880--2891. {PMLR}, July 2021.

\bibitem{mok2021AdvRushSearchingAdversarially}
Jisoo Mok, Byunggook Na, Hyeokjun Choe, and Sungroh Yoon.
\newblock {{AdvRush}}: {{Searching}} for {{Adversarially Robust Neural
  Architectures}}.
\newblock In {\em Proceedings of the {{IEEE}}/{{CVF International Conference}}
  on {{Computer Vision}}}, pages 12322--12332, 2021.

\bibitem{galloway2019BatchNormalizationCause}
Angus Galloway, Anna Golubeva, Thomas Tanay, Medhat Moussa, and Graham~W.
  Taylor.
\newblock Batch {{Normalization}} is a {{Cause}} of {{Adversarial
  Vulnerability}}.
\newblock {\em arXiv:1905.02161 [cs, stat]}, May 2019.

\bibitem{benz2021BatchNormalizationIncreases}
Philipp Benz, Chaoning Zhang, and In~So Kweon.
\newblock Batch {{Normalization Increases Adversarial Vulnerability}} and
  {{Decreases Adversarial Transferability}}: {{A Non-Robust Feature
  Perspective}}.
\newblock In {\em Proceedings of the {{IEEE}}/{{CVF International Conference}}
  on {{Computer Vision}}}, pages 7818--7827, 2021.

\bibitem{kong2022WhyDoesBatch}
Fei Kong, Fangqi Liu, Kaidi Xu, and Xiaoshuang Shi.
\newblock Why does batch normalization induce the model vulnerability on
  adversarial images?
\newblock {\em World Wide Web}, pages 1--19, July 2022.

\bibitem{zhang2021InterpretingImprovingAdversarial}
Chongzhi Zhang, Aishan Liu, Xianglong Liu, Yitao Xu, Hang Yu, Yuqing Ma, and
  Tianlin Li.
\newblock Interpreting and {{Improving Adversarial Robustness}} of {{Deep
  Neural Networks With Neuron Sensitivity}}.
\newblock {\em IEEE Transactions on Image Processing}, 30:1291--1304, 2021.

\bibitem{moraffah2020CausalInterpretabilityMachine}
Raha Moraffah, Mansooreh Karami, Ruocheng Guo, Adrienne Raglin, and Huan Liu.
\newblock Causal {{Interpretability}} for {{Machine Learning}} - {{Problems}},
  {{Methods}} and {{Evaluation}}.
\newblock {\em ACM SIGKDD Explorations Newsletter}, 22(1):18--33, May 2020.

\bibitem{narendra2018ExplainingDeepLearning}
Tanmayee Narendra, Anush Sankaran, Deepak Vijaykeerthy, and Senthil Mani.
\newblock Explaining {{Deep Learning Models}} using {{Causal Inference}}.
\newblock {\em arXiv:1811.04376 [cs, stat]}, November 2018.

\bibitem{janzing2020FeatureRelevanceQuantification}
Dominik Janzing, Lenon Minorics, and Patrick Bloebaum.
\newblock Feature relevance quantification in explainable {{AI}}: {{A}} causal
  problem.
\newblock In {\em International {{Conference}} on {{Artificial Intelligence}}
  and {{Statistics}}}, pages 2907--2916. {PMLR}, June 2020.

\bibitem{cifar10}
Alex Krizhevsky, Geoffrey Hinton, et~al.
\newblock Learning multiple layers of features from tiny images.
\newblock 2009.

\bibitem{imagenette}
Imagenette.
\newblock fast.ai, 2020.

\bibitem{deng2012mnist}
Li~Deng.
\newblock The mnist database of handwritten digit images for machine learning
  research.
\newblock {\em IEEE Signal Processing Magazine}, 29(6):141--142, 2012.

\bibitem{he2016DeepResidualLearning}
Kaiming He, Xiangyu Zhang, Shaoqing Ren, and Jian Sun.
\newblock Deep {{Residual Learning}} for {{Image Recognition}}.
\newblock In {\em 2016 {{IEEE Conference}} on {{Computer Vision}} and {{Pattern
  Recognition}} ({{CVPR}})}, pages 770--778, {Las Vegas, NV, USA}, June 2016.
  {IEEE}.

\bibitem{tan2021EfficientNetV2SmallerModels}
Mingxing Tan and Quoc~V. Le.
\newblock {{EfficientNetV2}}: {{Smaller Models}} and {{Faster Training}}, June
  2021.

\bibitem{carlini2017EvaluatingRobustnessNeural}
Nicholas Carlini and David Wagner.
\newblock Towards {{Evaluating}} the {{Robustness}} of {{Neural Networks}}.
\newblock In {\em 2017 {{IEEE Symposium}} on {{Security}} and {{Privacy}}
  ({{SP}})}, pages 39--57, May 2017.

\bibitem{chen2020HopSkipJumpAttackQueryEfficientDecisionBased}
Jianbo Chen, Michael~I. Jordan, and Martin~J. Wainwright.
\newblock {{HopSkipJumpAttack}}: {{A Query-Efficient Decision-Based Attack}}.
\newblock {\em arXiv:1904.02144 [cs, math, stat]}, April 2020.

\bibitem{abadi2016TensorflowSystemLargescale}
Mart{\'i}n Abadi, Paul Barham, Jianmin Chen, Zhifeng Chen, Andy Davis, Jeffrey
  Dean, Matthieu Devin, Sanjay Ghemawat, Geoffrey Irving, Michael Isard, et~al.
\newblock Tensorflow: {{A}} system for large-scale machine learning.
\newblock In {\em 12th \{\vphantom\}{{USENIX}}\vphantom\{\} Symposium on
  Operating Systems Design and Implementation
  (\{\vphantom\}{{OSDI}}\vphantom\{\} 16)}, pages 265--283, 2016.

\bibitem{art2018}
Maria-Irina Nicolae, Mathieu Sinn, Minh~Ngoc Tran, Beat Buesser, Ambrish Rawat,
  Martin Wistuba, Valentina Zantedeschi, Nathalie Baracaldo, Bryant Chen, Heiko
  Ludwig, Ian Molloy, and Ben Edwards.
\newblock Adversarial robustness toolbox v1.2.0.
\newblock {\em CoRR}, 1807.01069, 2018.

\end{thebibliography}

\clearpage
\appendix

\renewcommand{\thefigure}{A\arabic{figure}}
\setcounter{figure}{0}

\renewcommand{\theequation}{A\arabic{equation}}
\setcounter{equation}{0}

\renewcommand{\thetable}{A\arabic{table}}
\setcounter{table}{0}


\section{Statistical significance}
\label{apx:wrst}

To evaluate the statistical significance of the results, the Wilcoxon Rank-Sum test (WRST) has been employed to verify that the effects caused on the model by channel $j$ are significantly different than the effects caused by channel $k$. This is especially useful to prove that there are differences in the effects caused by the top-ranking channels (most vulnerable) and bottom-ranking channels (least vulnerable).

To do so, the WRST has been applied to compare the Average Effect on Logits (AEL) caused by pairs of channels. For values of $\gamma > 2$ there exist multiple combinations $\Phi$ where both channels $i$ and $j$ are present. To avoid the confounding effect that may arise by perturbing both channels at the same time, the WRST was applied between the subsets of channel combinations that exclusively include either channel, but not both. Using the notation defined above we define

\begin{align}
	\Omega^\gamma_{j\setminus k} = \Omega^\gamma_{j} \setminus (\Omega^\gamma_{j} \cap \Omega^\gamma_{k}) \\
	\textbf{AEL}^\gamma_{j\setminus k} = \{ AEA_\Phi : \Phi \in \Omega^\gamma_{j\setminus k} \}  
\end{align}

It follows that, for a pair of channels $j$ and $k$, the WRST has been applied between the sets $\textbf{AEL}^\gamma_{j\setminus k}$ and $\textbf{AEL}^\gamma_{k\setminus j}$.

The results of the tests of statistical significance are visualized as an heatmap reporting the p-value associated with the WRST between two channels for a fixed $gamma$ (e.g. Figure \ref{apx:fig:cifar:gamma3}(d)). The values in the colorbar are limited to the interval [0, 0.10] for an easier comparison between plots; hence all the values beyond 0.10 are clipped to the usual p-value threshold of $0.10$. In the heatmaps, a cell with a red color indicates that the effects caused by the two channels are statistically significant with $p < 0.05$, whereas a blue indicates a value $p > 0.05$.

\section{Additional Plots}
\label{apx:plots}

We report here additional plots for the results obtained in the experiments.

\begin{figure}
	\centering
	\begin{subfigure}{0.45\linewidth}
		\includegraphics[width=\linewidth]{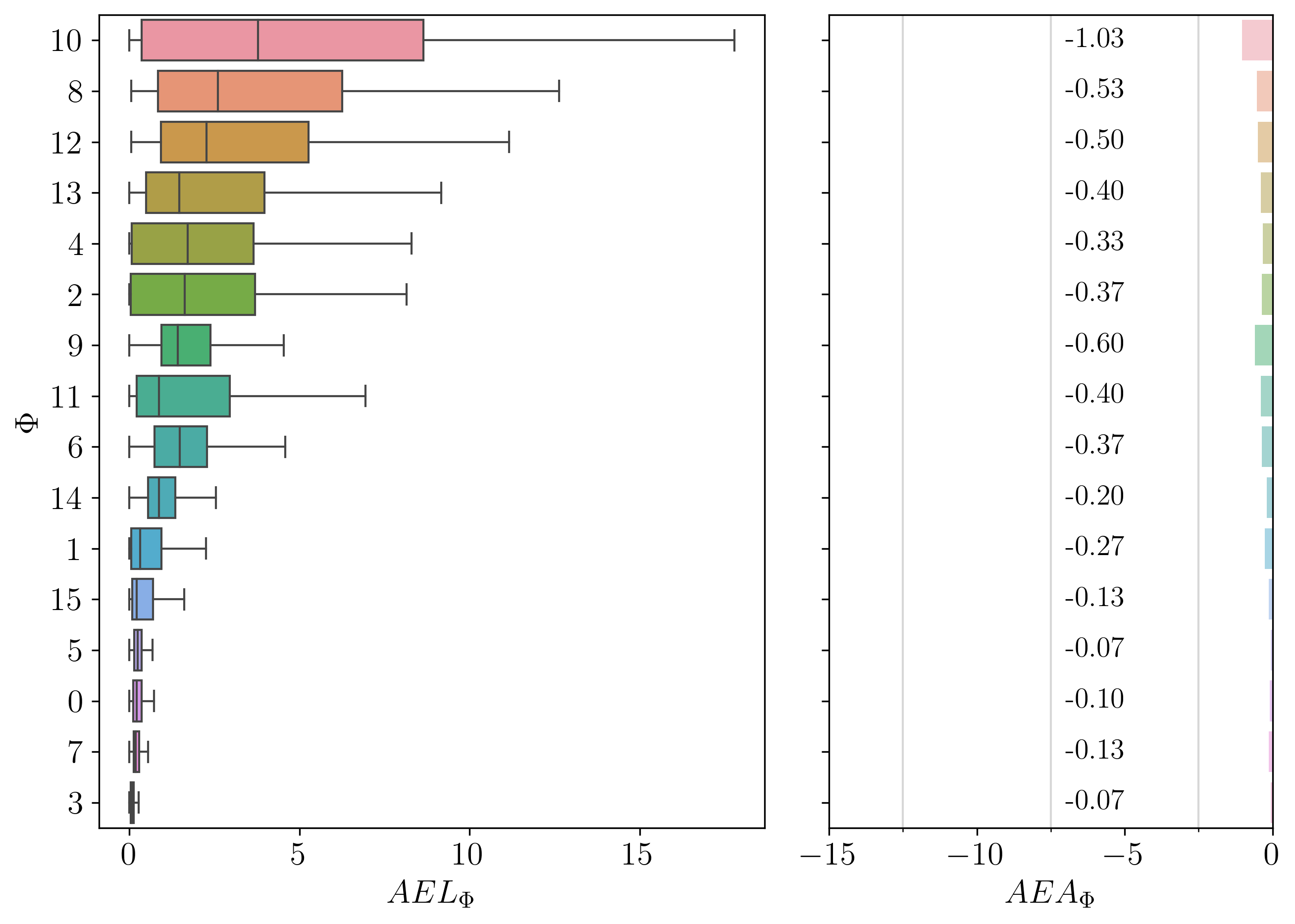}
		\caption{}
	\end{subfigure}
	\begin{subfigure}{0.45\linewidth}
		\includegraphics[width=\linewidth]{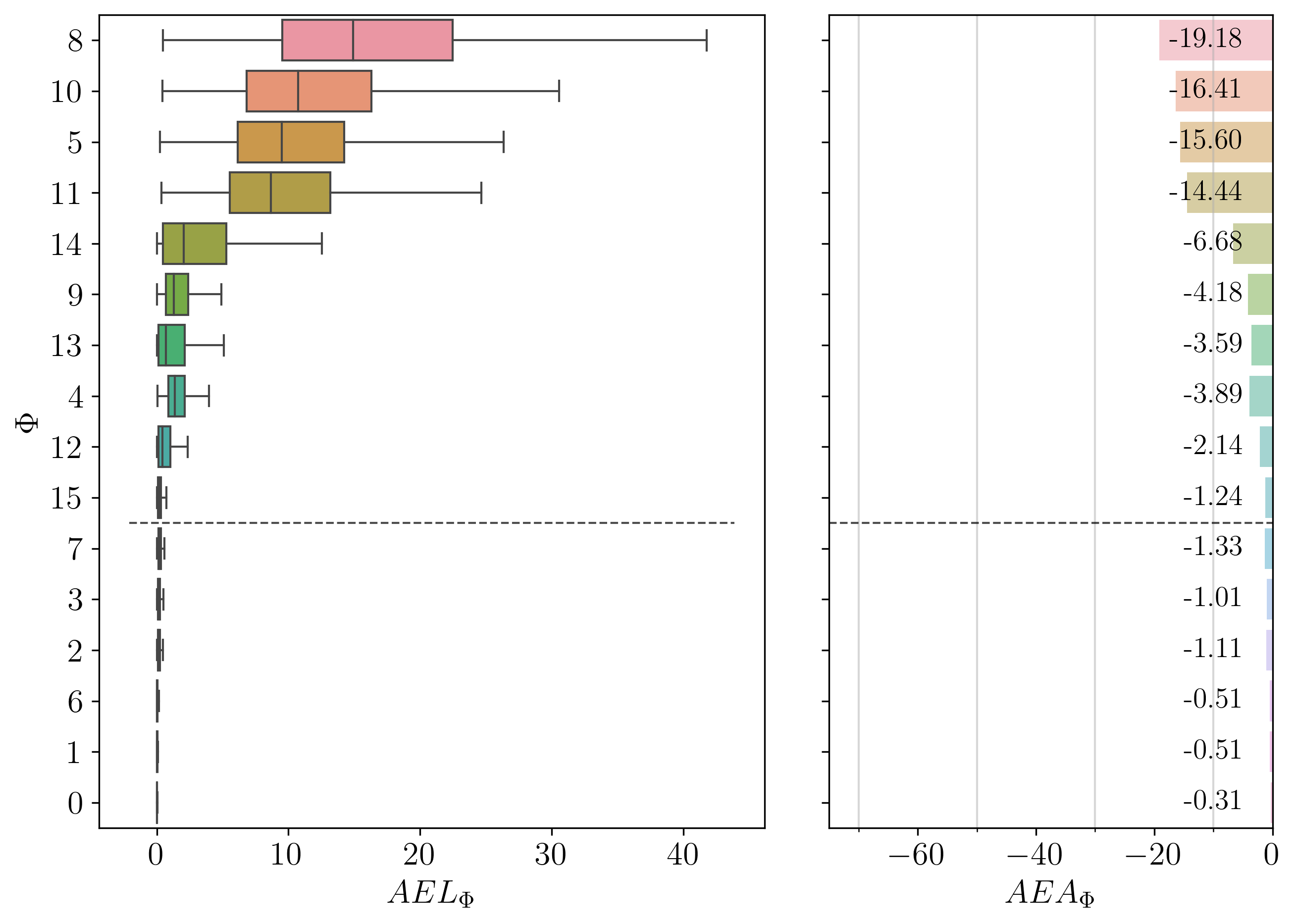}
		\caption{}
	\end{subfigure}
	\\
	\begin{subfigure}{0.45\linewidth}
		\includegraphics[width=\linewidth]{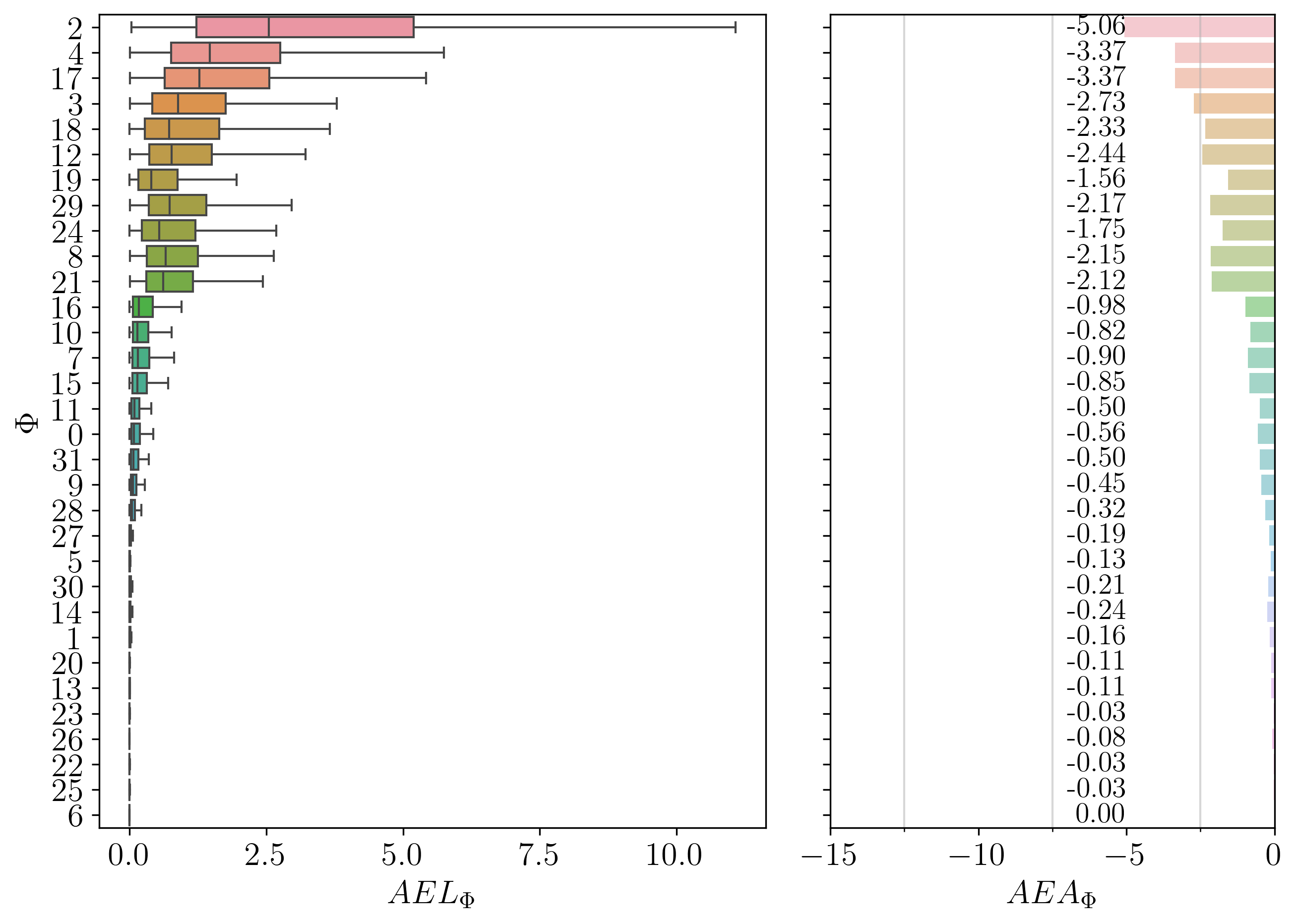}
		\caption{}
	\end{subfigure}
	\caption{Effects of Adversarial Intervention with Auto-PGD on the examined models for $\gamma=1$. (a) MNIST-37; (b) CIFAR-10; (c) Imagenette.}
	\label{fig:gamma1}
\end{figure}

Figure \ref{fig:gamma1} shows the effects of applying \textit{Adversarial Intervention} (Auto-PGD) with $\gamma=1$; i.e. by applying an adversarial perturbation only on a single channel of the learned features. The plots reveal that for the more complex datasets like CIFAR-10 (b) and Imagenette (c), a perturbation on a single channel is sufficient to generate big disruptions to the model. The same does not happen in the MNIST-37 case, likely due to the simplicity of the data and the redundancy of the learned features.

Figure \ref{apx:fig:mnist:gamma7} reports the plots related to \textit{Adversarial Intervention} (Auto-PGD) applied to the MNIST-37 model with $\gamma=7$. Certain combinations of 7 channels or more affect the model by reducing its accuracy by more than 70\%,

Figure \ref{apx:fig:cifar:gamma3} includes two additional plots to the results already shown in Figure \ref{fig:cifar:gamma3} with respect to the \textit{Adversarial Intervention} (Auto-PGD) on the CIFAR-10 model with $\gamma=3$. The histogram in (c) shows the distribution of the effects caused by all the combinations of $\gamma=3$ channels. The plot reveals how the majority of 3-channel-combinations have marginal effects on the model, whereas there exist only a handful of combinations that strongly disrupt the behavior of the model. The heatmap of p-values in (d) shows that the average effects caused by the top-ranking channels are significantly different than the bottom-ranking ones.
As expected, with a higher number of channels that are perturbed through \textit{Adversarial Intervention}, more channel combinations have disruptive effects on the model.
We report the effects caused by \textit{Adversarial Intervention} with $\gamma=7$ in Figure \ref{apx:fig:cifar:gamma7}. Here, the majority of combinations have big effects on the model output, however there are still combinations that cause negligible effects on the model.
With more channels involved in the intervention, it becomes harder to isolate the effects of a single channel. Hence, the differences in the single-channel effects are less significant for this value of $\gamma$.

In the following pages, we report the results obtained by applying \textit{Adversarial Intervention} to the CIFAR-10 model with the two other attacks: Figures \ref{apx:fig:cifar:gamma3:cwl2}-\ref{apx:fig:cifar:gamma14:cwl2} for Carlini \& Wagner (C\&W); Figures \ref{apx:fig:cifar:gamma3:hsj}-\ref{apx:fig:cifar:gamma14:hsj} for Hop-Skip Jump (HSJ).
As discussed in the main manuscript, the plots show that these attacks have a different mechanism of affecting the model as the perturbations appear to be more spread among the channels. It follows that more channels need to be perturbed at the same time through Adversarial Intervention to achieve greater disruptions to the model. 
Lastly, we report the correlation plots between the $AEL_\Phi$ (Average Effect on Logits) obtained by the same channel combinations across the 3 different attacks and 3 distinct values of $\gamma$. These plots show that the high positive correlation in the effects caused by perturbing the same channels is consistent among the 3 attacks that have been investigated.

\newcommand{\tpl}{0.45}

\newcommand{\fourfigdesc}{(a) Top and bottom 10 $\Phi$ combinations, ranked by $AEL_\Phi$; (b) Channel-wise effects on logits and accuracy; (c) Histogram of the effects on accuracy caused by all the possible combinations of $\gamma$ channels; (d) Heatmap of p-values of double-tailed pairwise Wilcoxon Rank-Sum test on the $AEL$. For clarity, the outlier points beyond 1.5x the inter-quartile range have been excluded in the boxplots of (a) and (b).}


\begin{figure*}
	\newcommand{\spl}{0.45}
	\centering
	\begin{subfigure}{\spl\linewidth}
		\includegraphics[width=\linewidth]{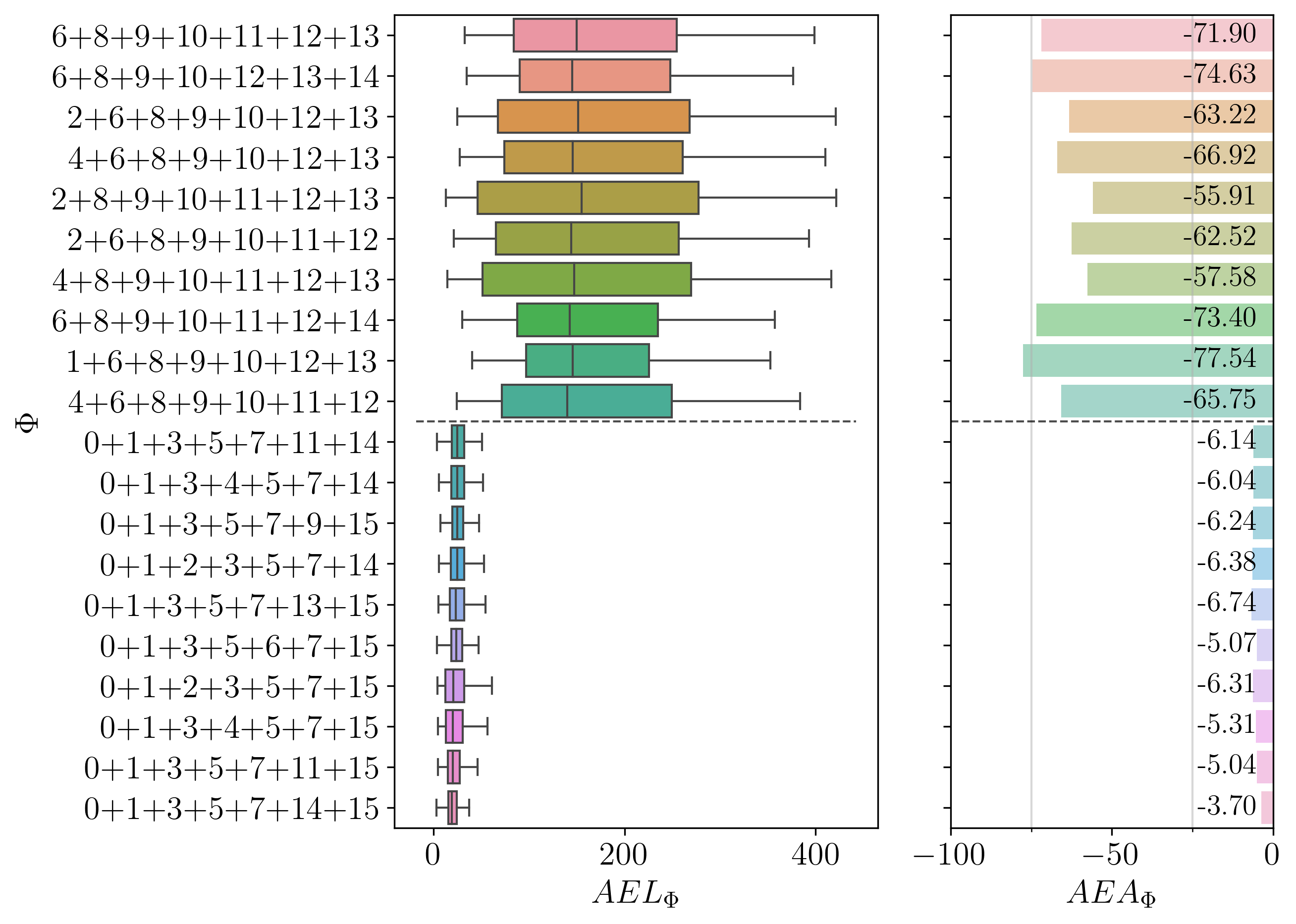}
		\caption{}
	\end{subfigure}
	\quad
	\begin{subfigure}{\spl\linewidth}
		\includegraphics[width=\linewidth]{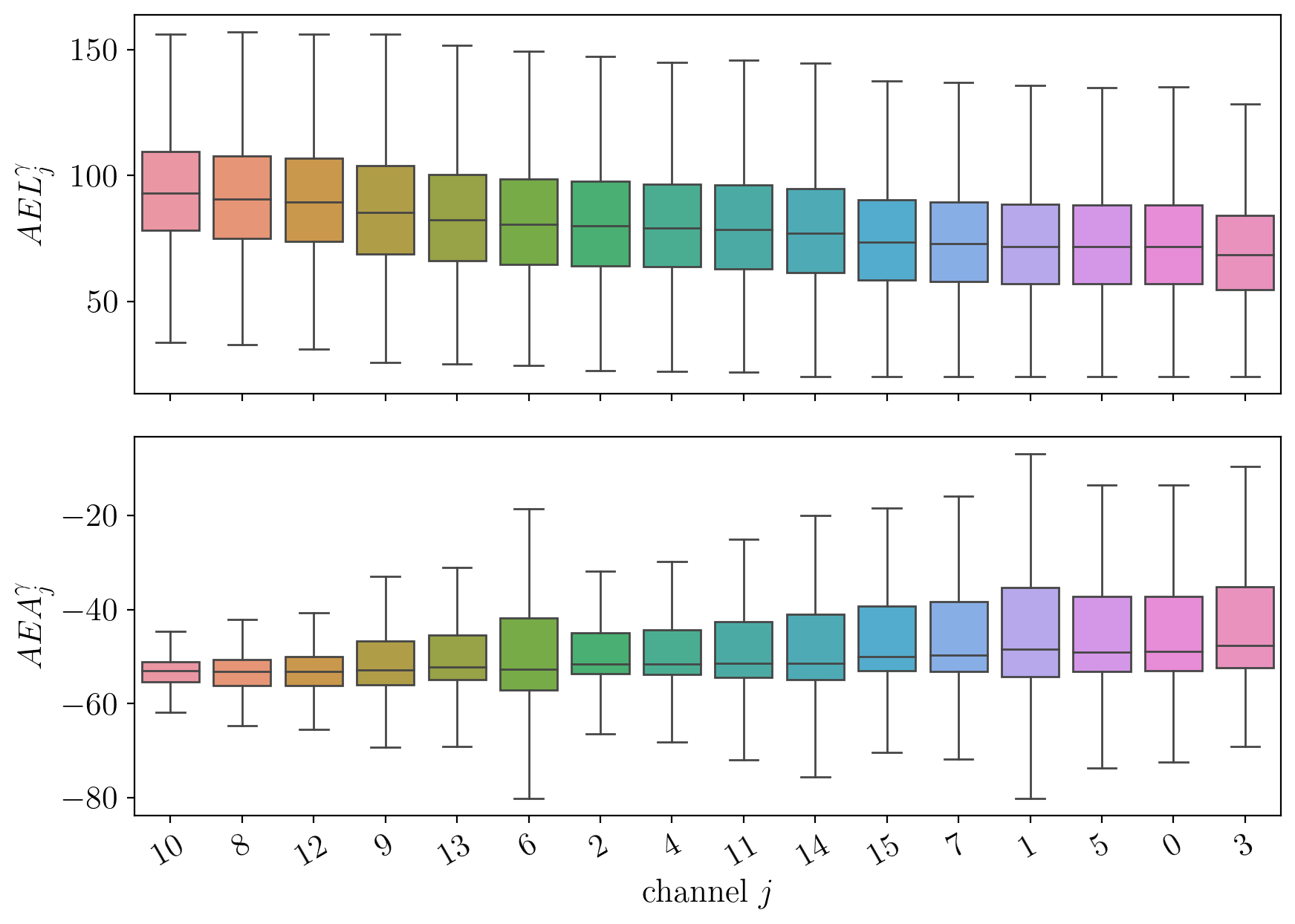}
		\caption{}
	\end{subfigure}
	\\
	\begin{subfigure}{\spl\linewidth}
		\includegraphics[width=\linewidth]{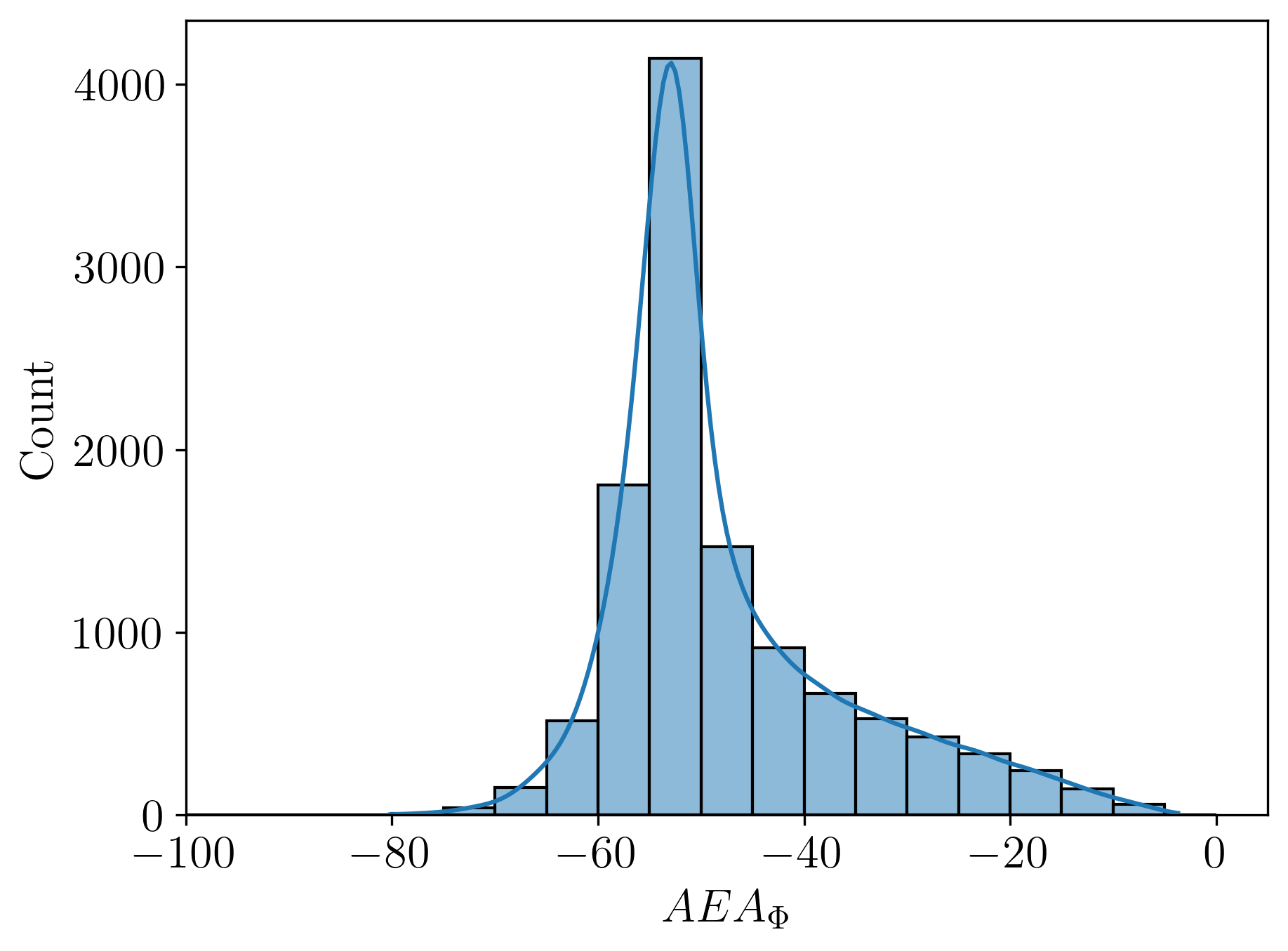}
		\caption{}
	\end{subfigure}
	\quad
	\begin{subfigure}{\spl\linewidth}
		\includegraphics[width=\linewidth]{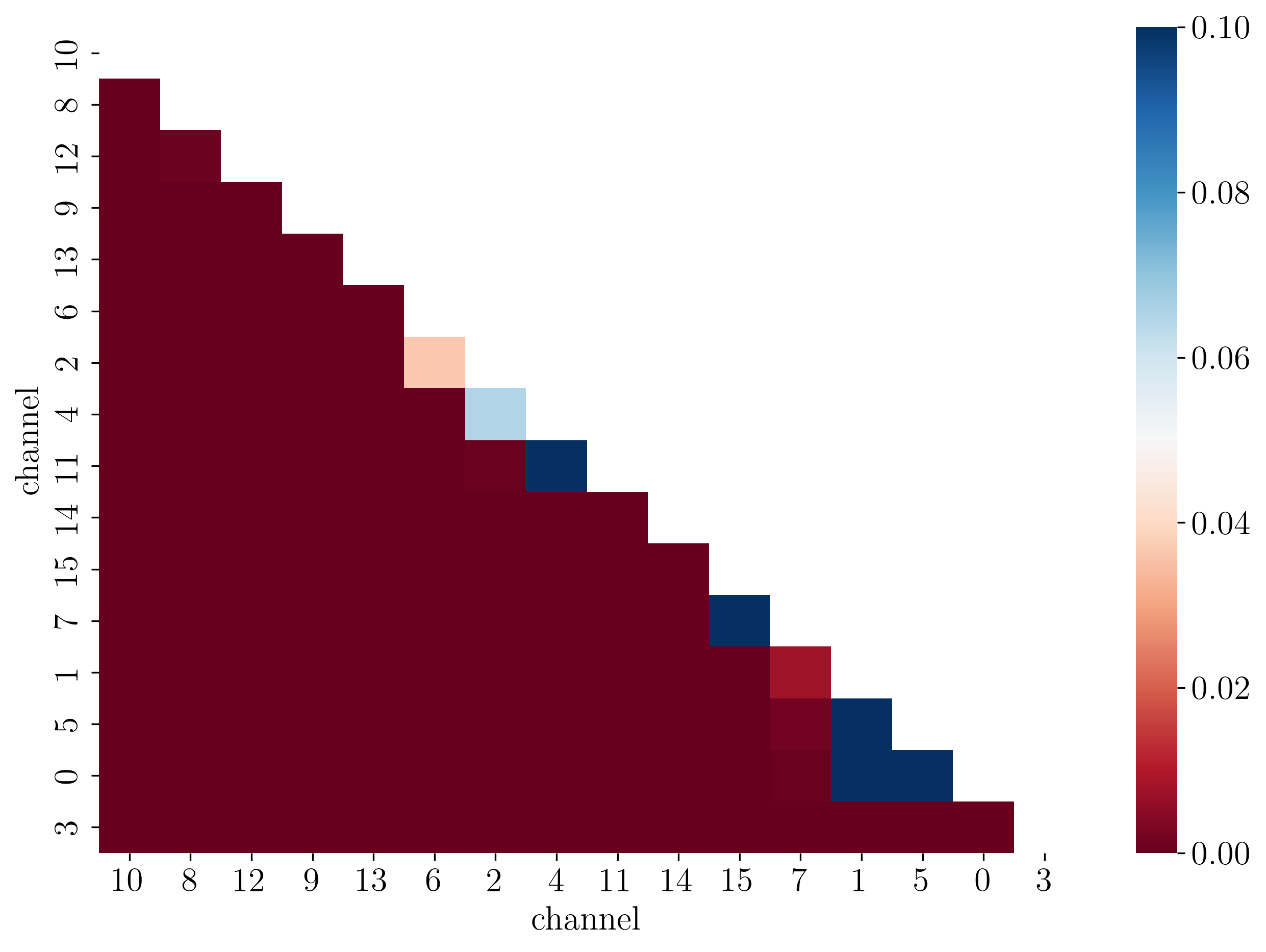}
		\caption{}
	\end{subfigure}
	\caption{Results of \textit{Adversarial Intervention} (Auto-PGD) on MNIST-37 model for $\gamma=7$. \fourfigdesc}
	\label{apx:fig:mnist:gamma7}
\end{figure*}


\begin{figure*}
	\newcommand{\spl}{0.45}
	\centering
	\begin{subfigure}{\spl\linewidth}
		\includegraphics[width=\linewidth]{cifar10-o3.png}
		\caption{}
	\end{subfigure}
	\quad
	\begin{subfigure}{\spl\linewidth}
		\includegraphics[width=\linewidth]{cifar10-o3_sc.png}
		\caption{}
	\end{subfigure}
	\\
	\begin{subfigure}{\spl\linewidth}
		\includegraphics[width=\linewidth]{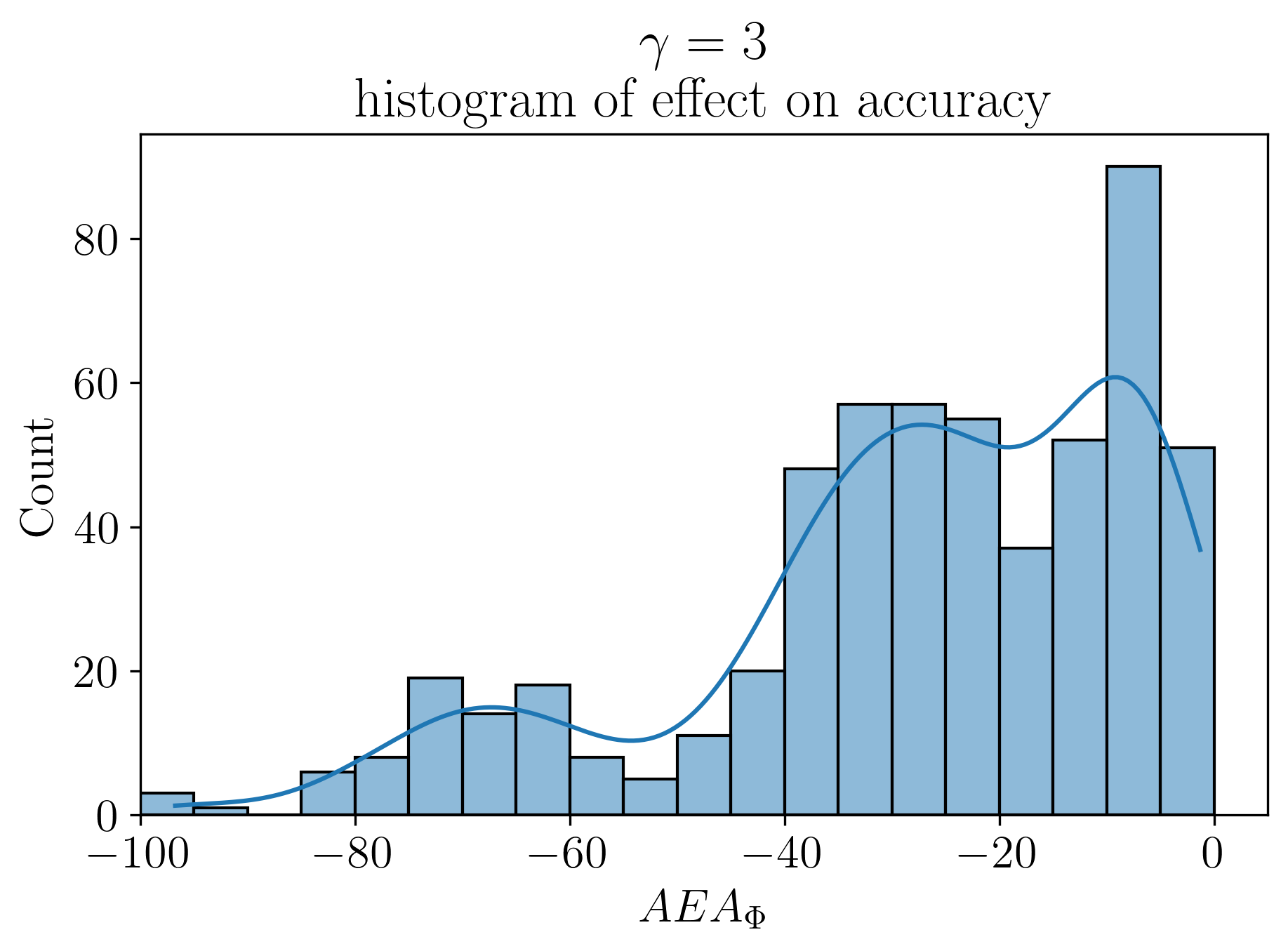}
		\caption{}
	\end{subfigure}
	\quad
	\begin{subfigure}{\spl\linewidth}
		\includegraphics[width=\linewidth]{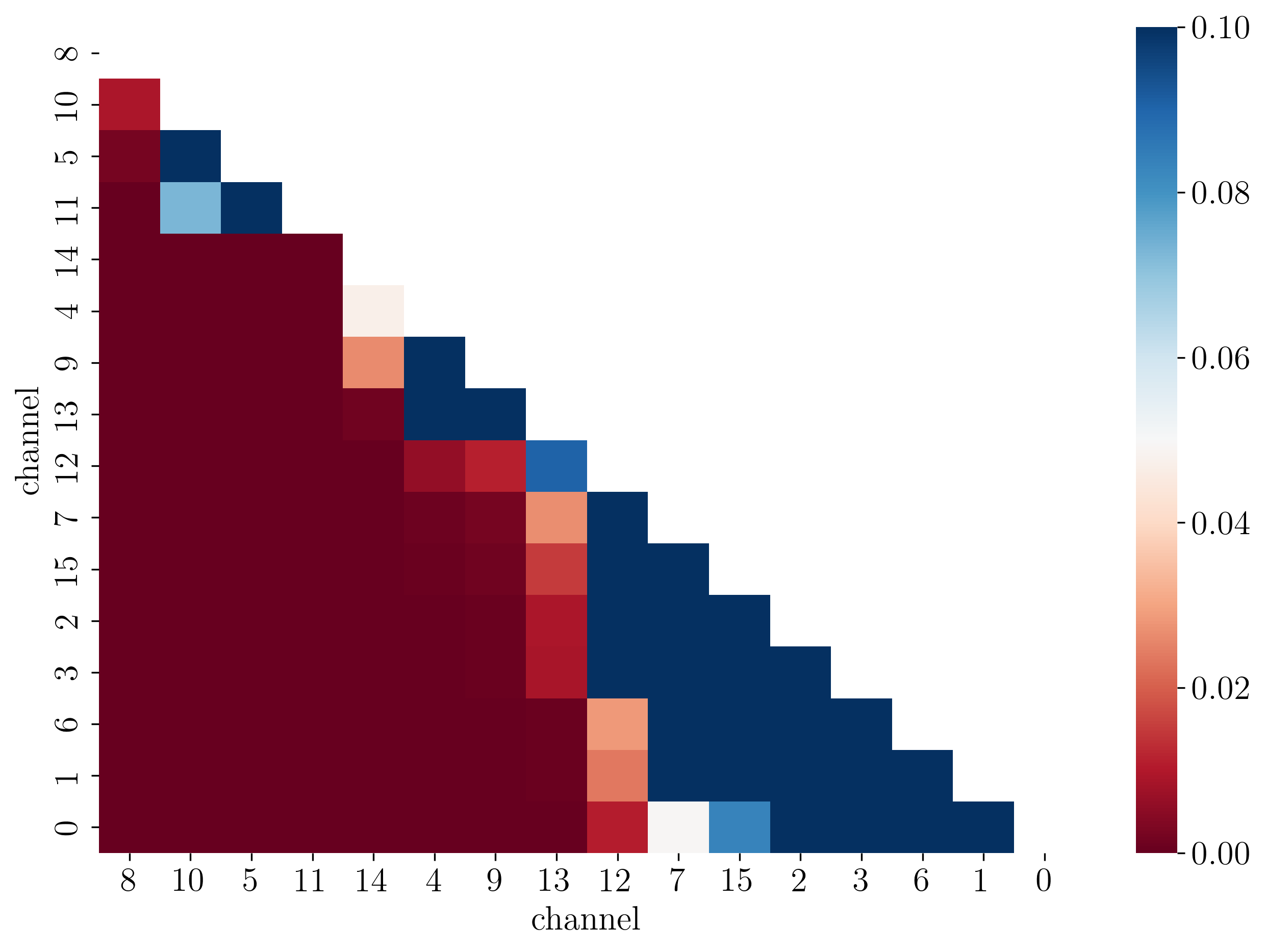}
		\caption{}
	\end{subfigure}
	\caption{Results of \textit{Adversarial Intervention} (Auto-PGD) on CIFAR-10 model for $\gamma=3$. \fourfigdesc}
	\label{apx:fig:cifar:gamma3}
\end{figure*}

\begin{figure*}
	\newcommand{\spl}{0.45}
	\centering
	\begin{subfigure}{\spl\linewidth}
		\includegraphics[width=\linewidth]{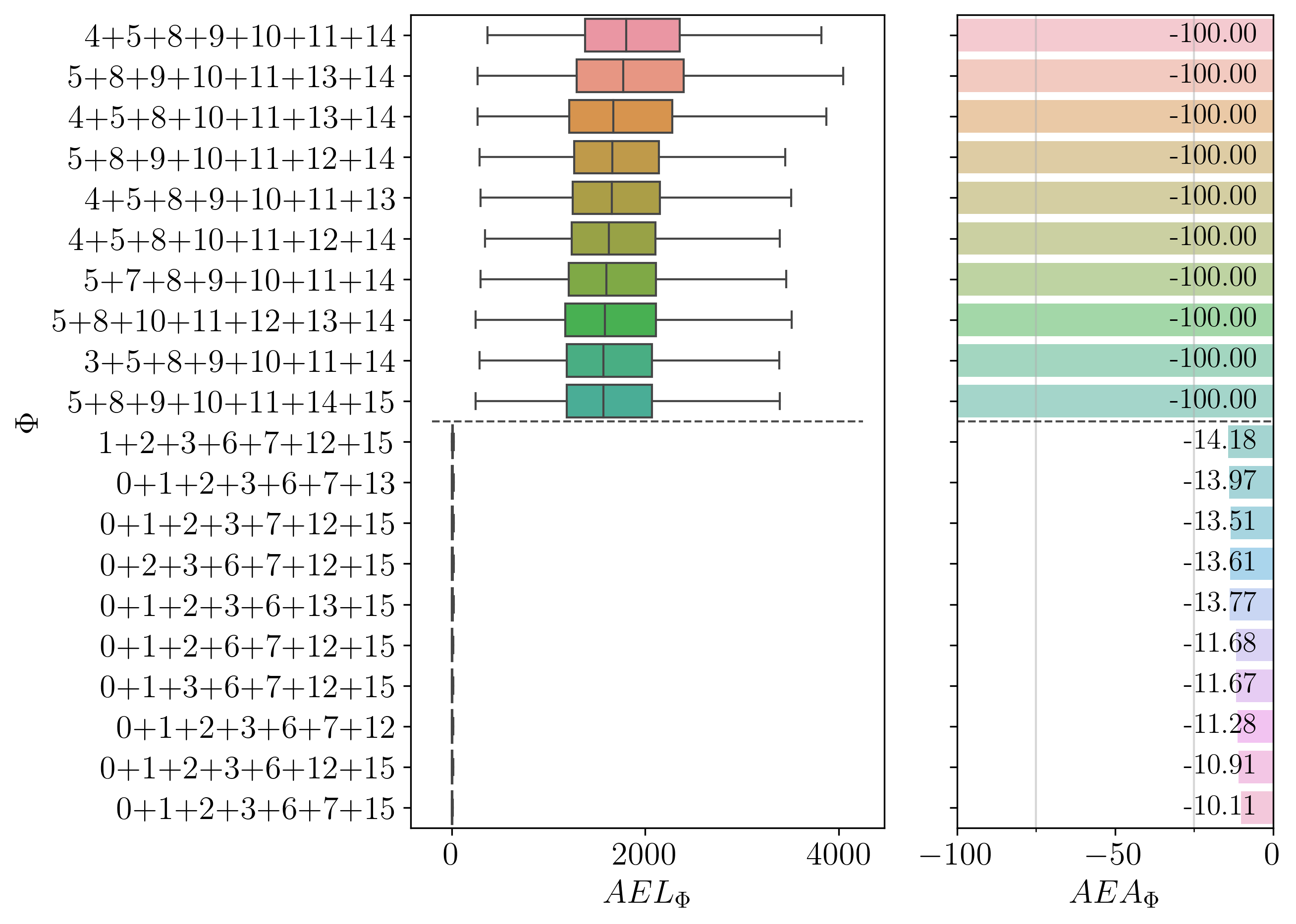}
		\caption{}
	\end{subfigure}
	\quad
	\begin{subfigure}{\spl\linewidth}
		\includegraphics[width=\linewidth]{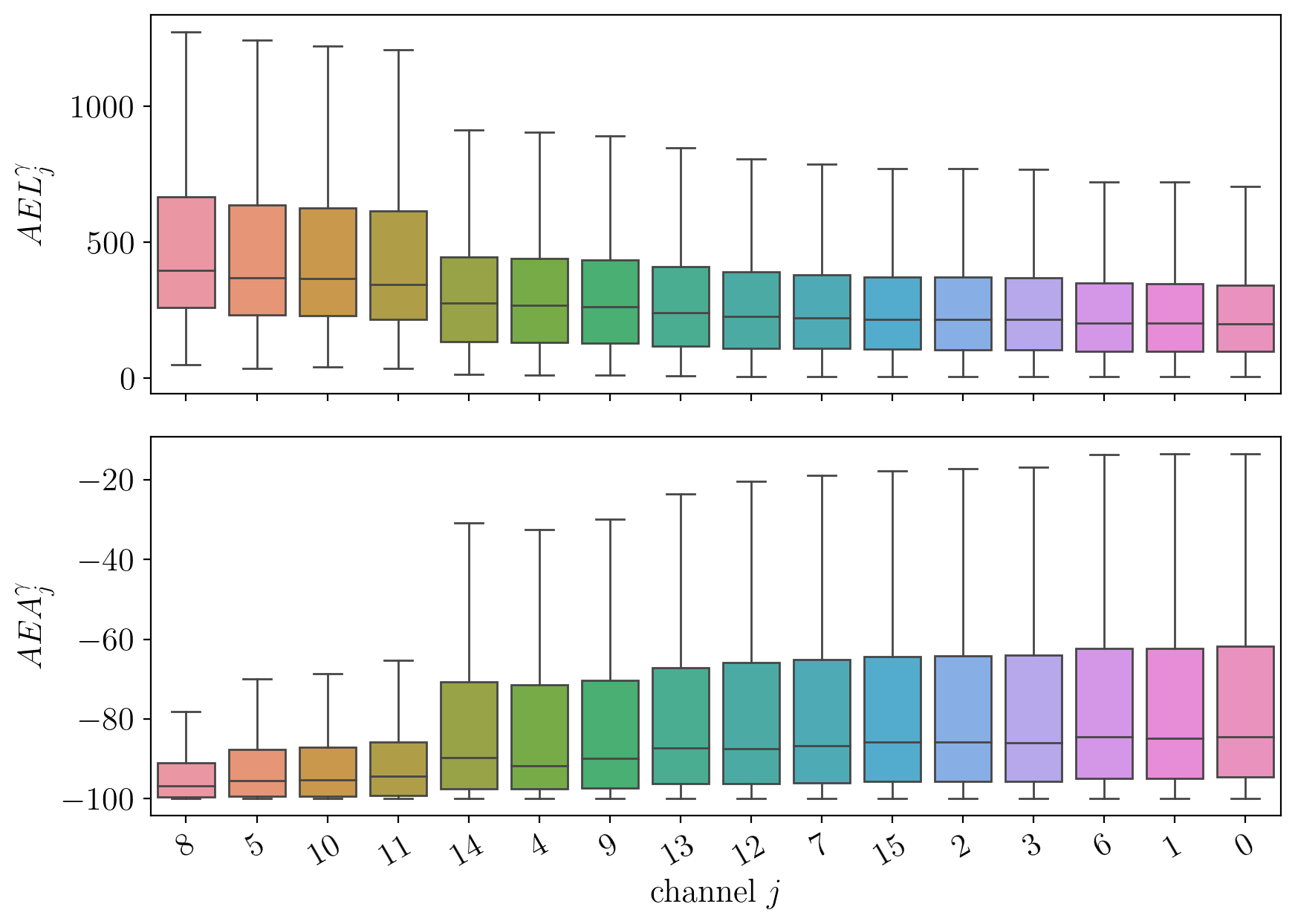}
		\caption{}
	\end{subfigure}
	\\
	\begin{subfigure}{\spl\linewidth}
		\includegraphics[width=\linewidth]{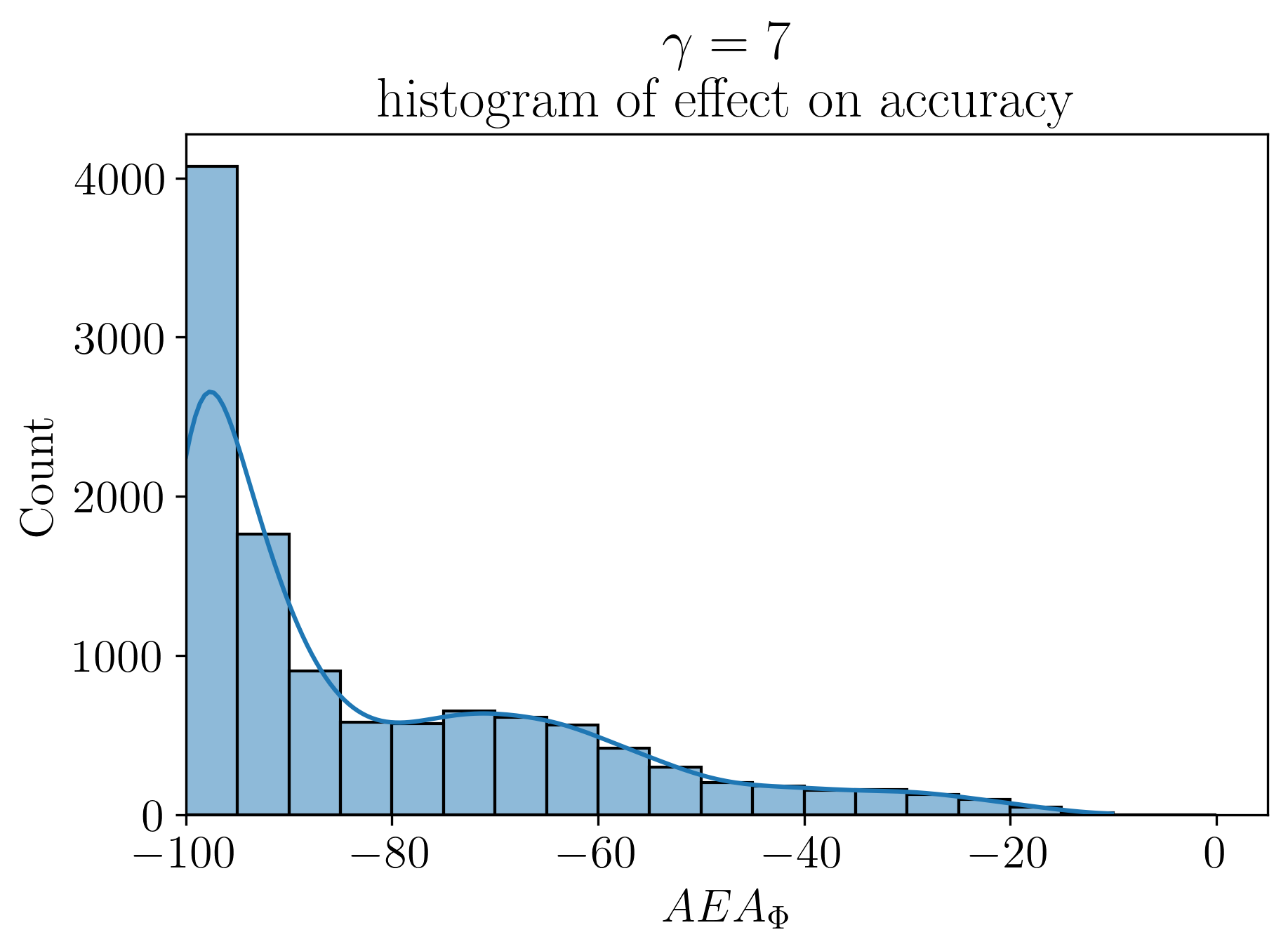}
		\caption{}
	\end{subfigure}
	\quad
	\begin{subfigure}{\spl\linewidth}
		\includegraphics[width=\linewidth]{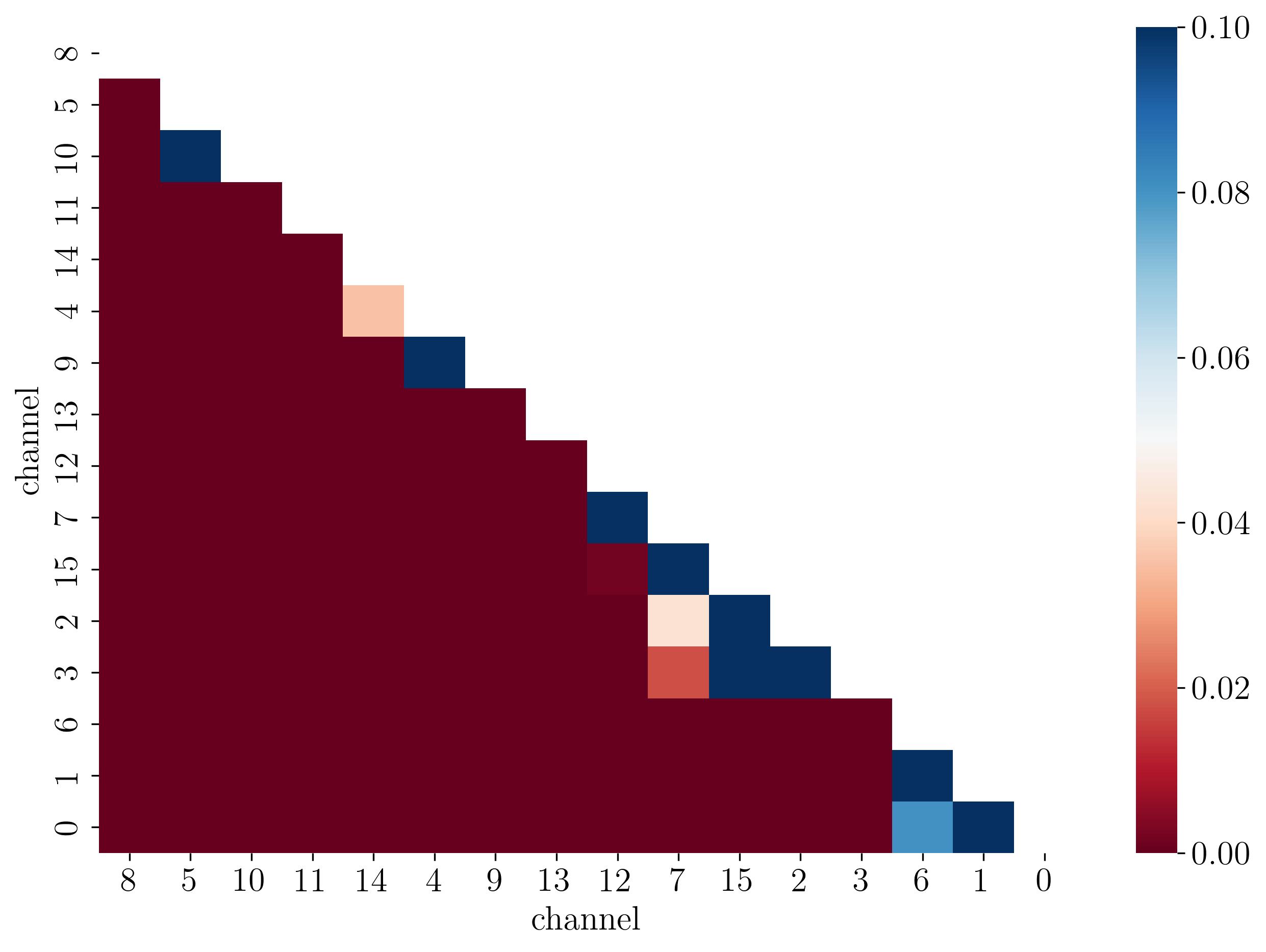}
		\caption{}
	\end{subfigure}
	\caption{Results of \textit{Adversarial Intervention} (Auto-PGD) on CIFAR-10 model for $\gamma=7$. \fourfigdesc}
	\label{apx:fig:cifar:gamma7}
\end{figure*}

\begin{figure*}
	\newcommand{\spl}{0.45}
	\centering
	\begin{subfigure}{\spl\linewidth}
		\includegraphics[width=\linewidth]{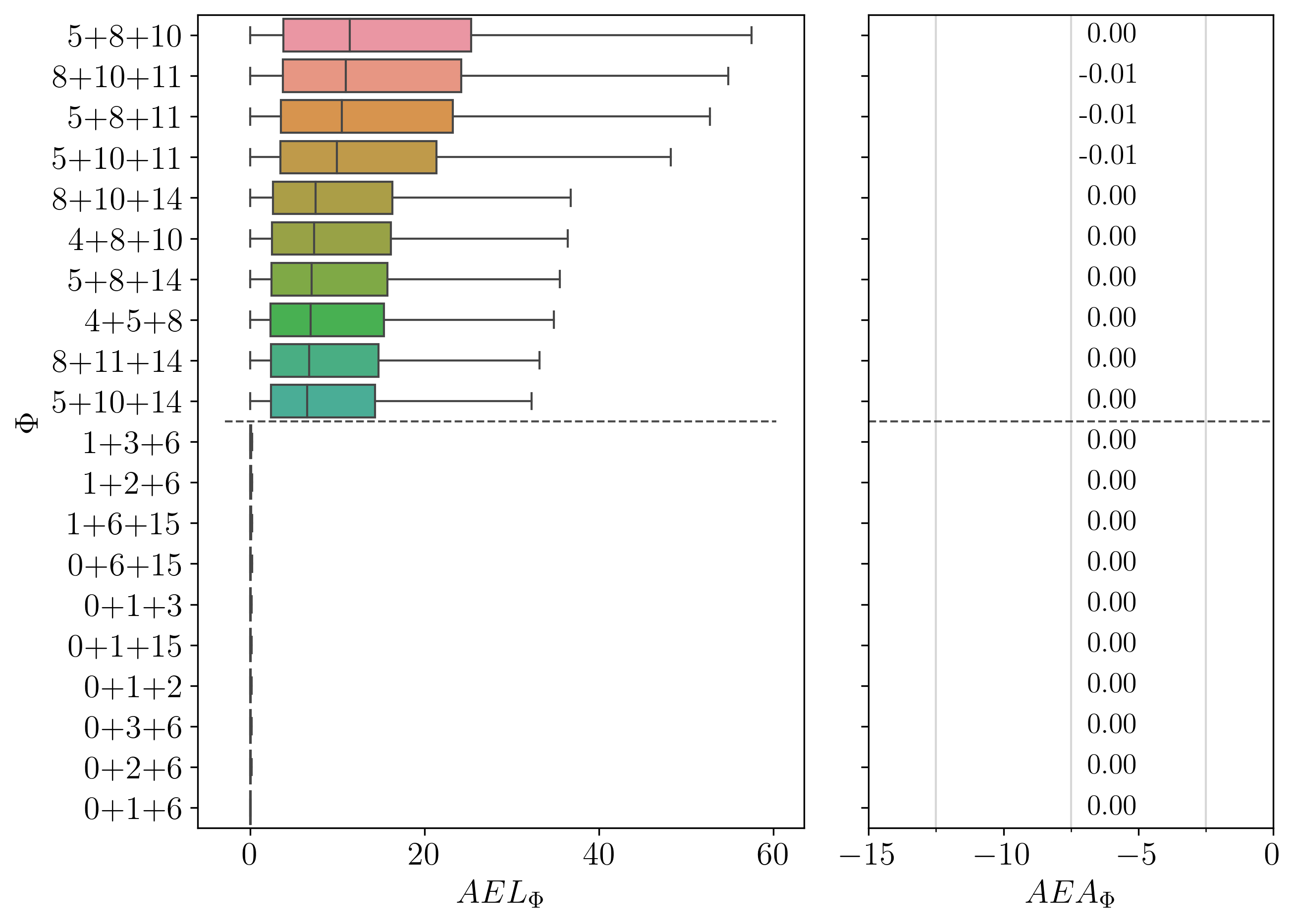}
		\caption{}
	\end{subfigure}
	\quad
	\begin{subfigure}{\spl\linewidth}
		\includegraphics[width=\linewidth]{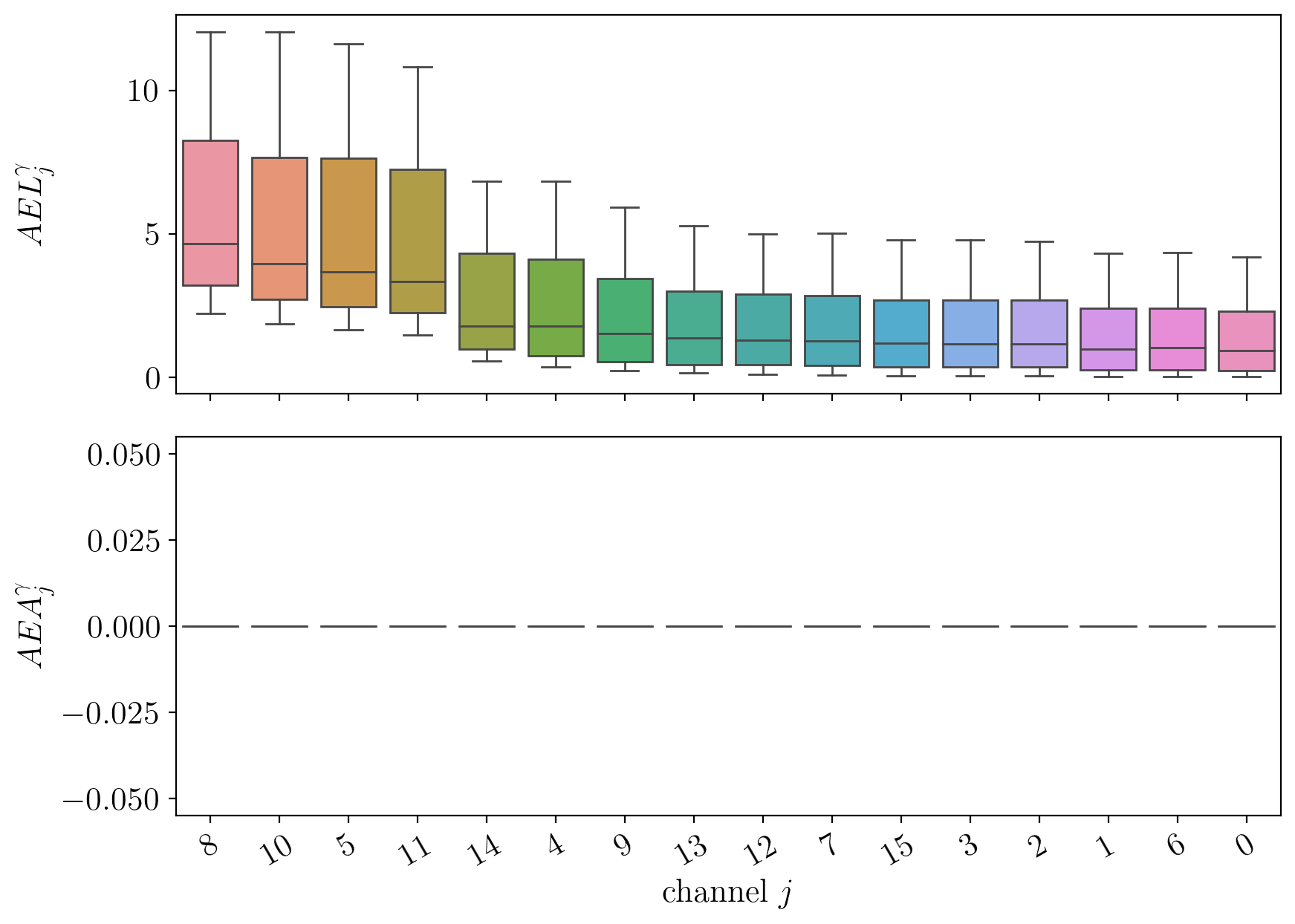}
		\caption{}
	\end{subfigure}
	\\
	\begin{subfigure}{\spl\linewidth}
		\includegraphics[width=\linewidth]{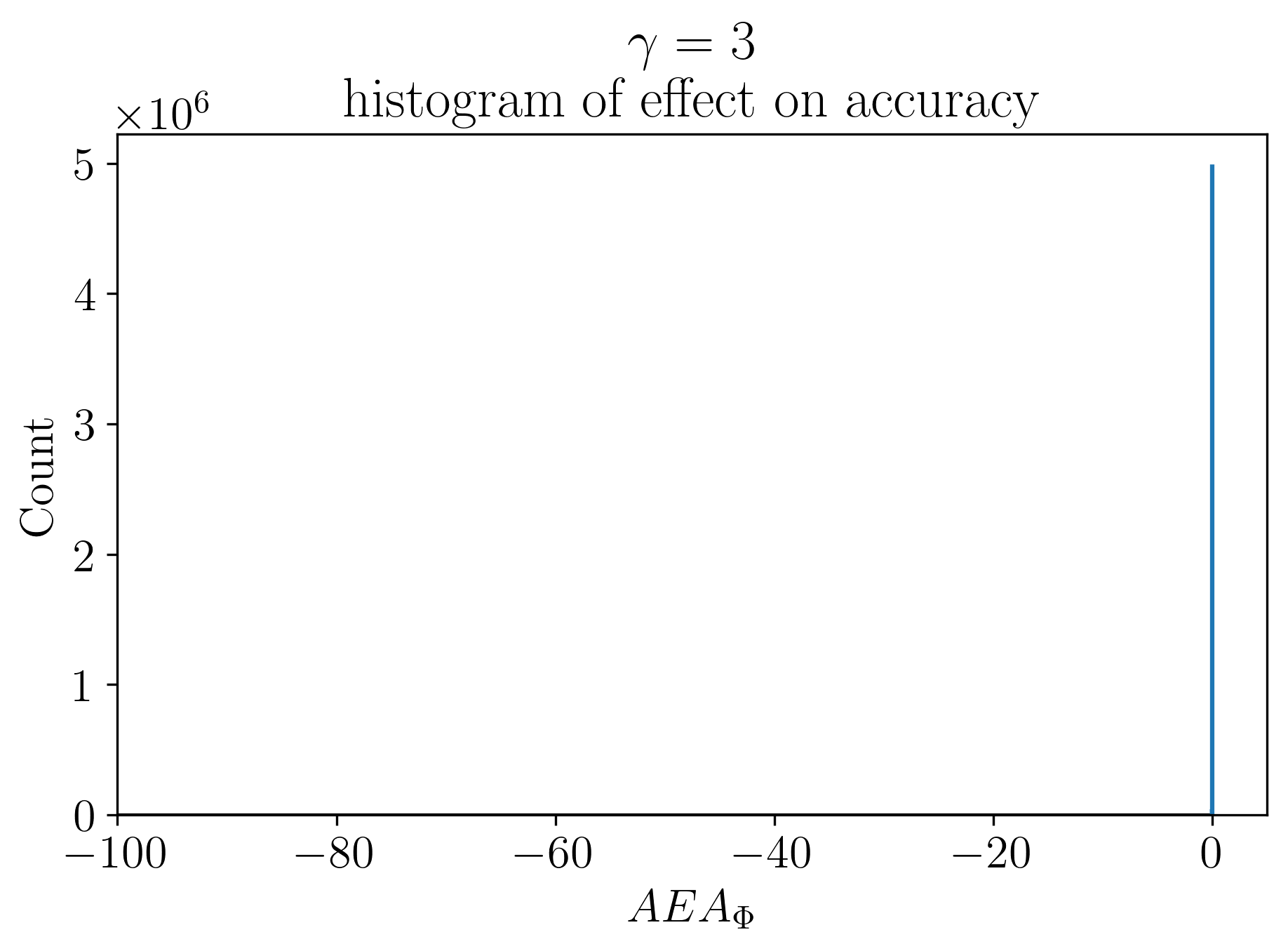}
		\caption{}
	\end{subfigure}
	\quad
	\begin{subfigure}{\spl\linewidth}
		\includegraphics[width=\linewidth]{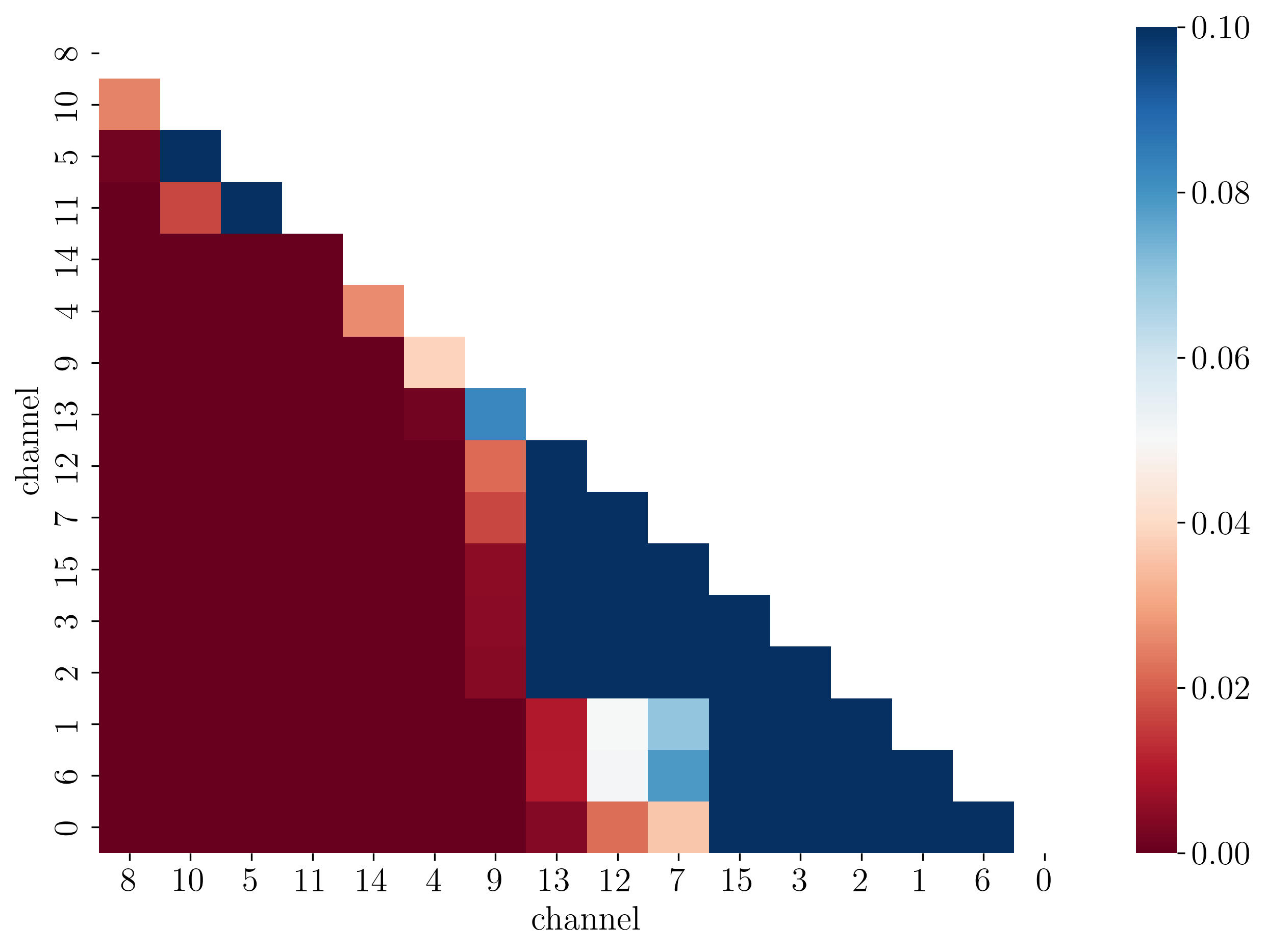}
		\caption{}
	\end{subfigure}
	\caption{Results of \textit{Adversarial Intervention} (Carlini \& Wagner) on CIFAR-10 model for $\gamma=3$. \fourfigdesc}
	\label{apx:fig:cifar:gamma3:cwl2}
\end{figure*}

\begin{figure*}
	\newcommand{\spl}{0.45}
	\centering
	\begin{subfigure}{\spl\linewidth}
		\includegraphics[width=\linewidth]{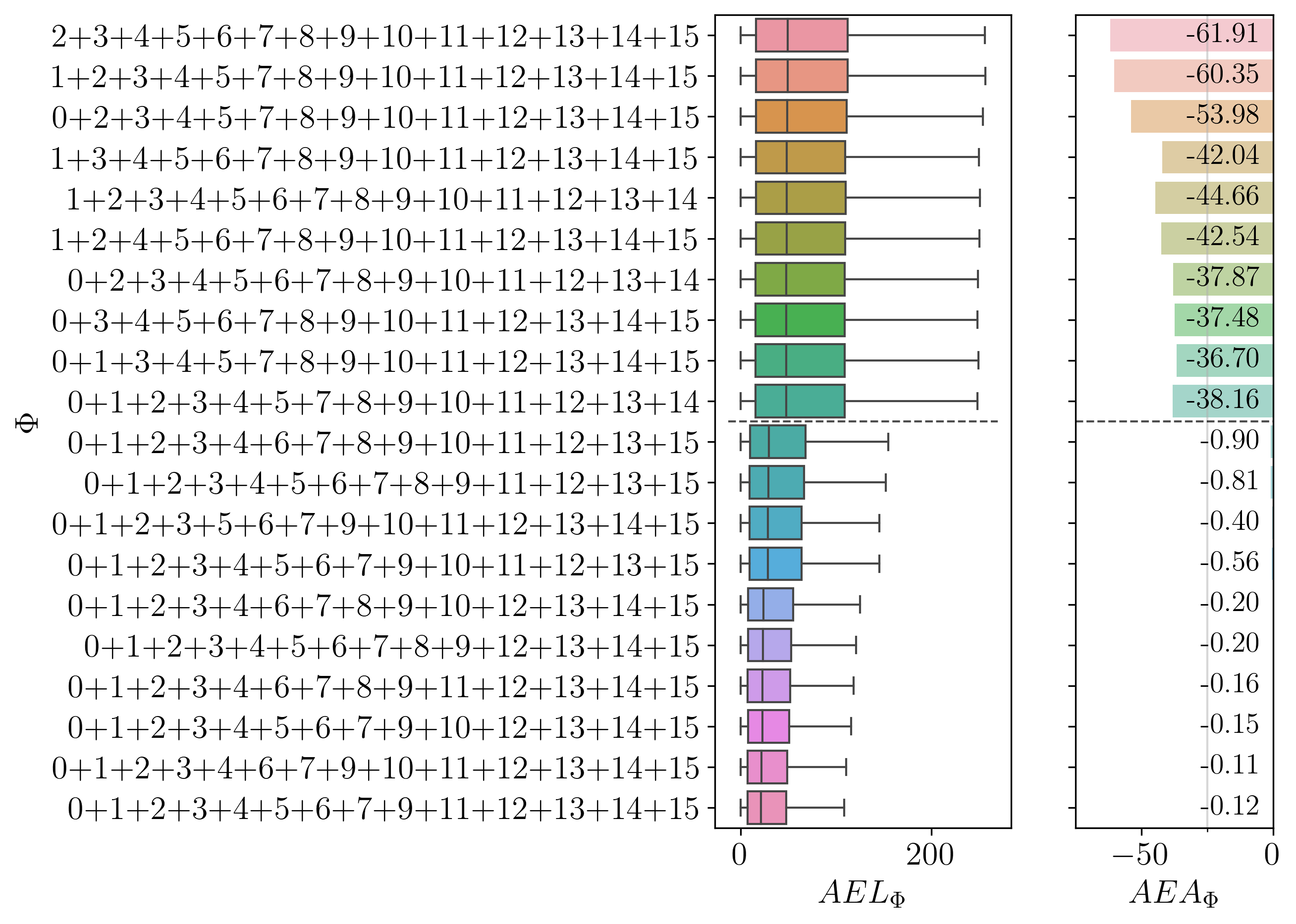}
		\caption{}
	\end{subfigure}
	\quad
	\begin{subfigure}{\spl\linewidth}
		\includegraphics[width=\linewidth]{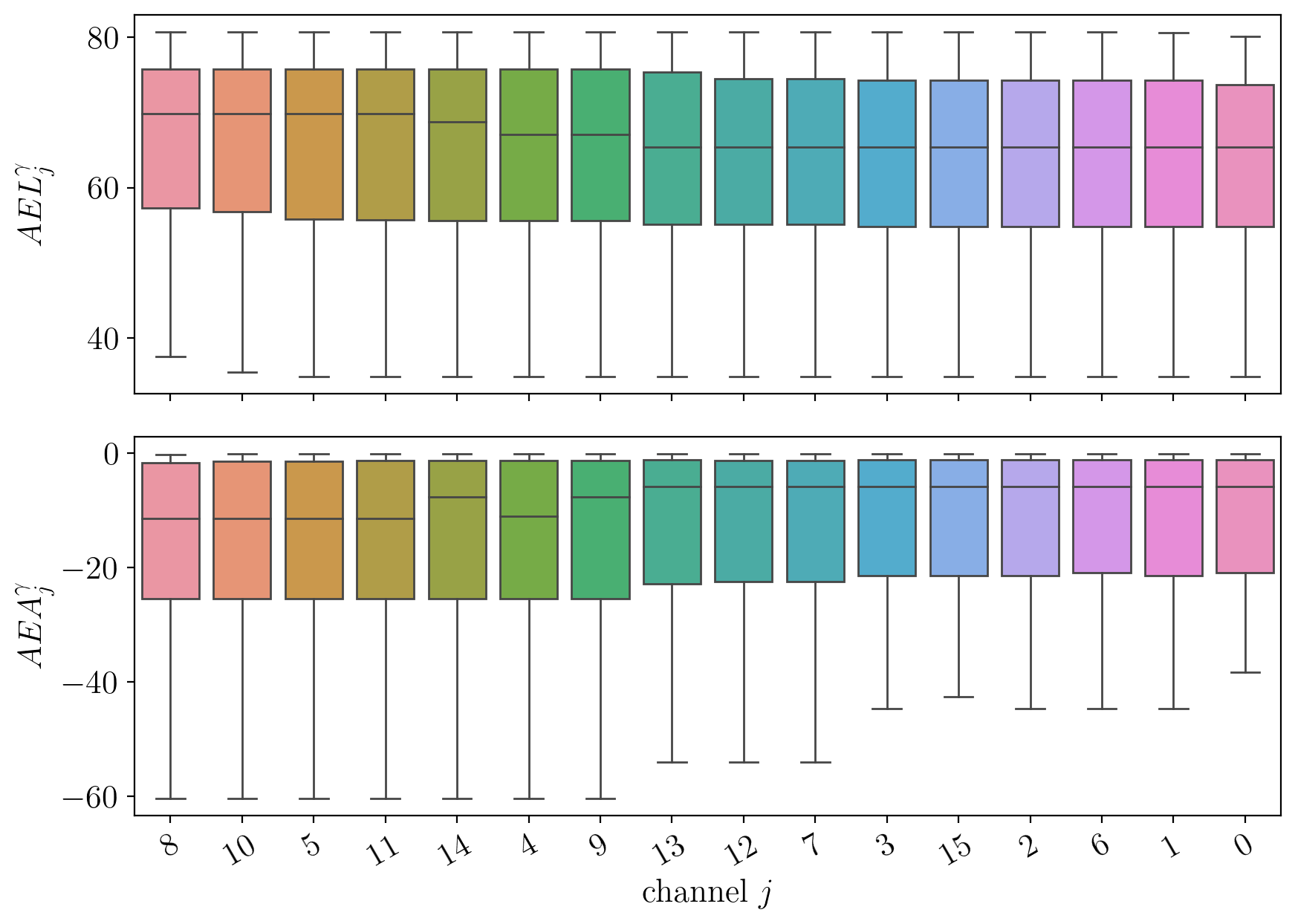}
		\caption{}
	\end{subfigure}
	\\
	\begin{subfigure}{\spl\linewidth}
		\includegraphics[width=\linewidth]{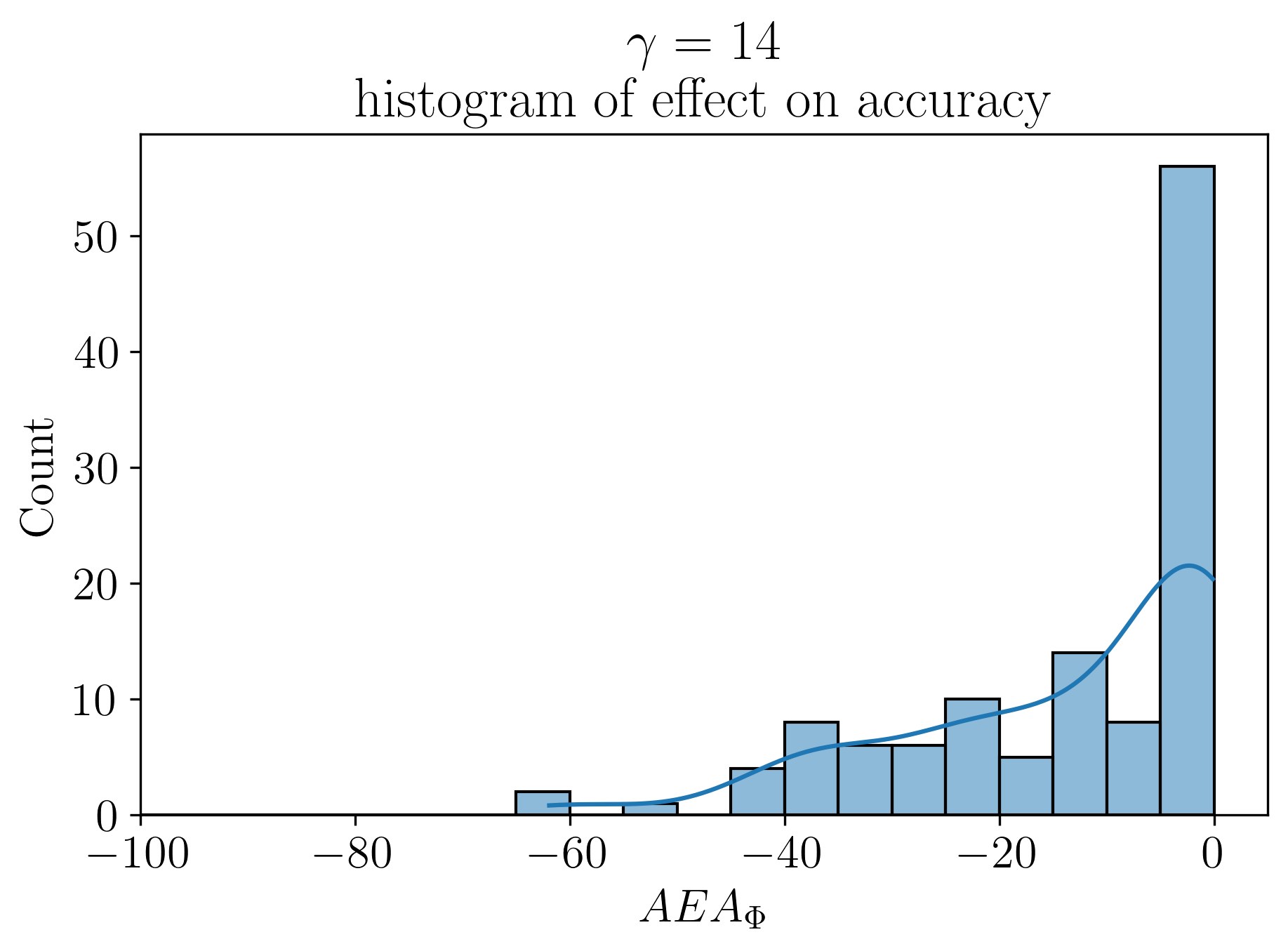}
		\caption{}
	\end{subfigure}
	\quad
	\begin{subfigure}{\spl\linewidth}
		\includegraphics[width=\linewidth]{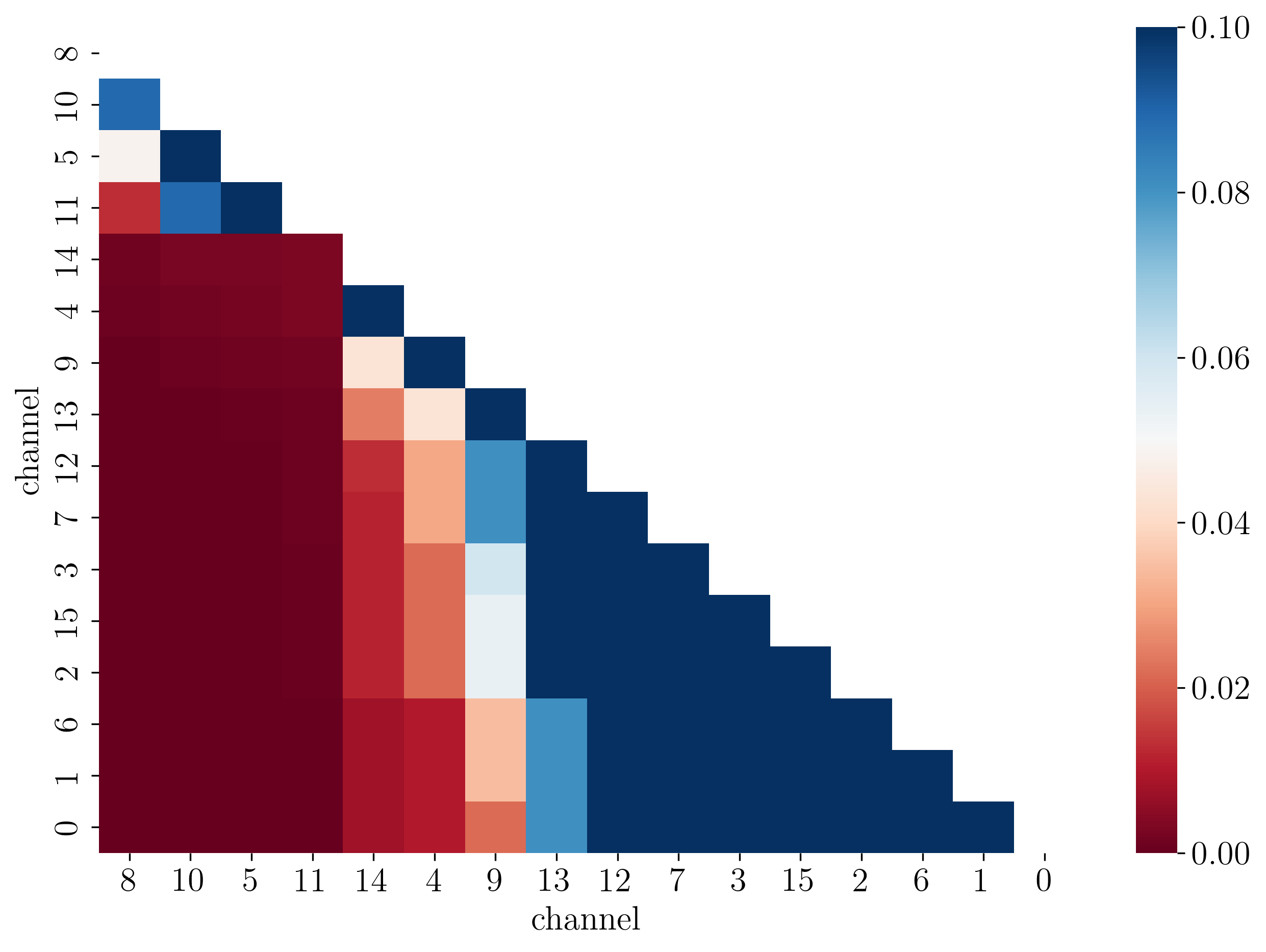}
		\caption{}
	\end{subfigure}
	\caption{Results of \textit{Adversarial Intervention} (Carlini \& Wagner) on CIFAR-10 model for $\gamma=14$. \fourfigdesc}
	\label{apx:fig:cifar:gamma14:cwl2}
\end{figure*}


\begin{figure*}
	\newcommand{\spl}{0.45}
	\centering
	\begin{subfigure}{\spl\linewidth}
		\includegraphics[width=\linewidth]{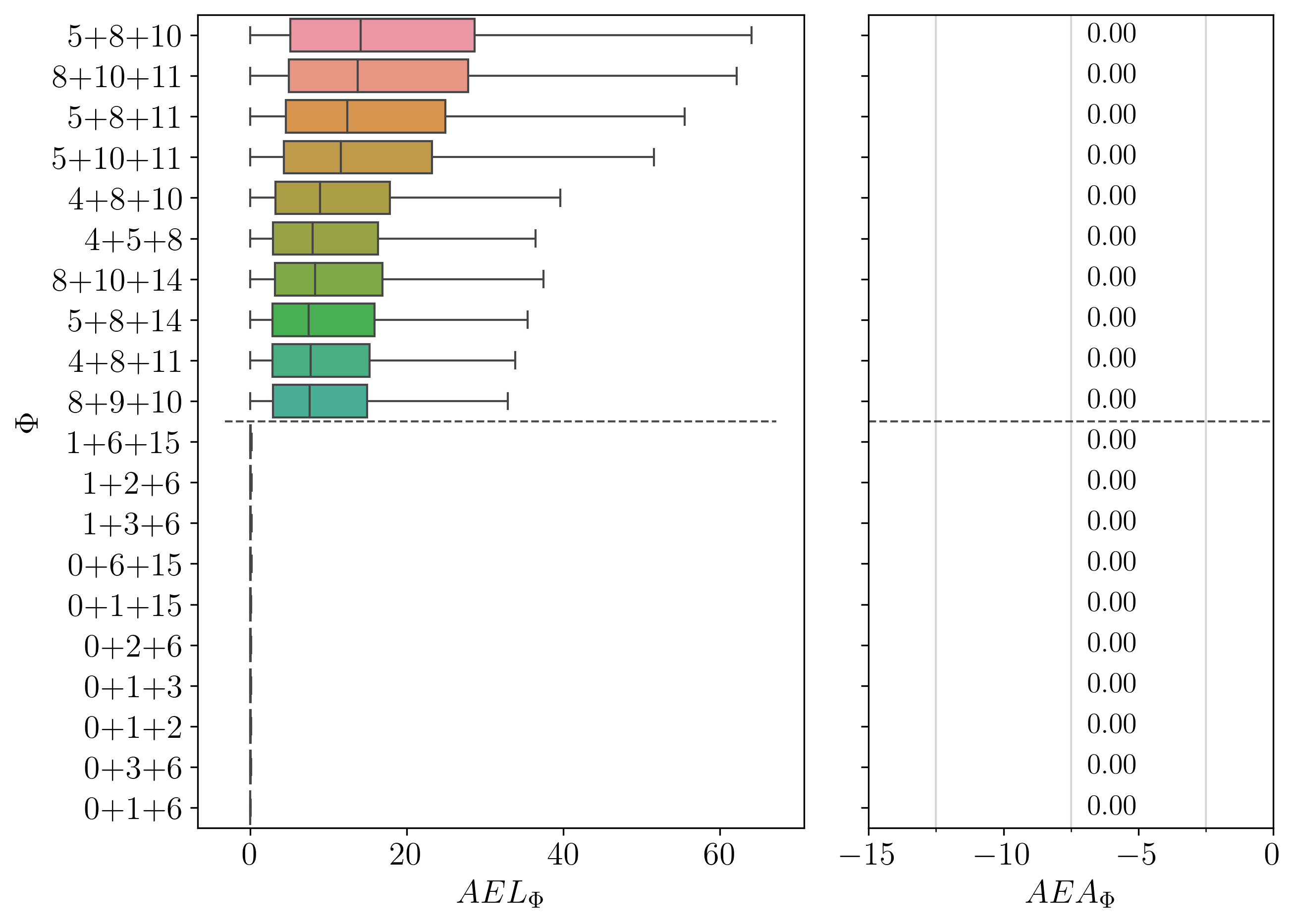}
		\caption{}
	\end{subfigure}
	\quad
	\begin{subfigure}{\spl\linewidth}
		\includegraphics[width=\linewidth]{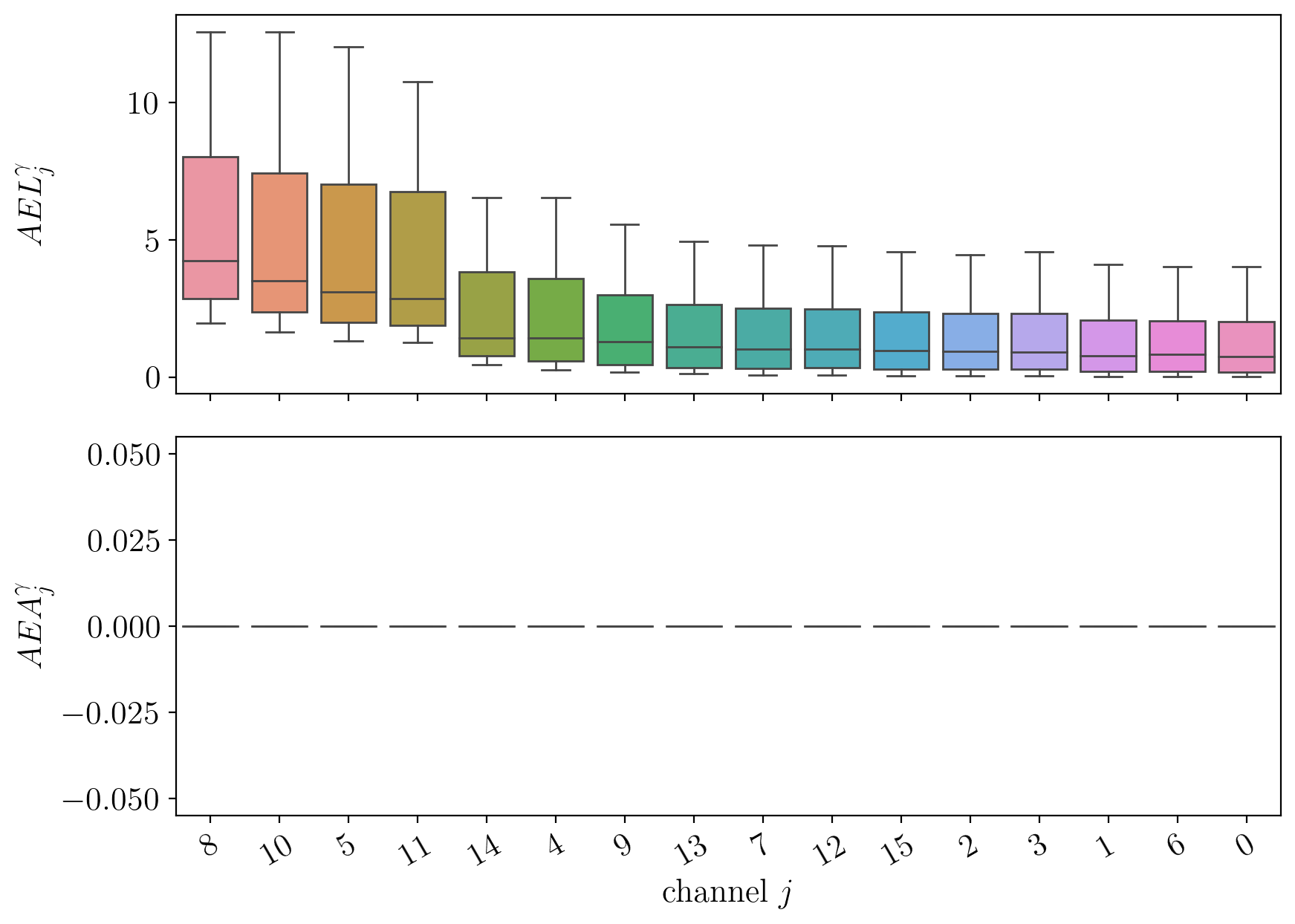}
		\caption{}
	\end{subfigure}
	\\
	\begin{subfigure}{\spl\linewidth}
		\includegraphics[width=\linewidth]{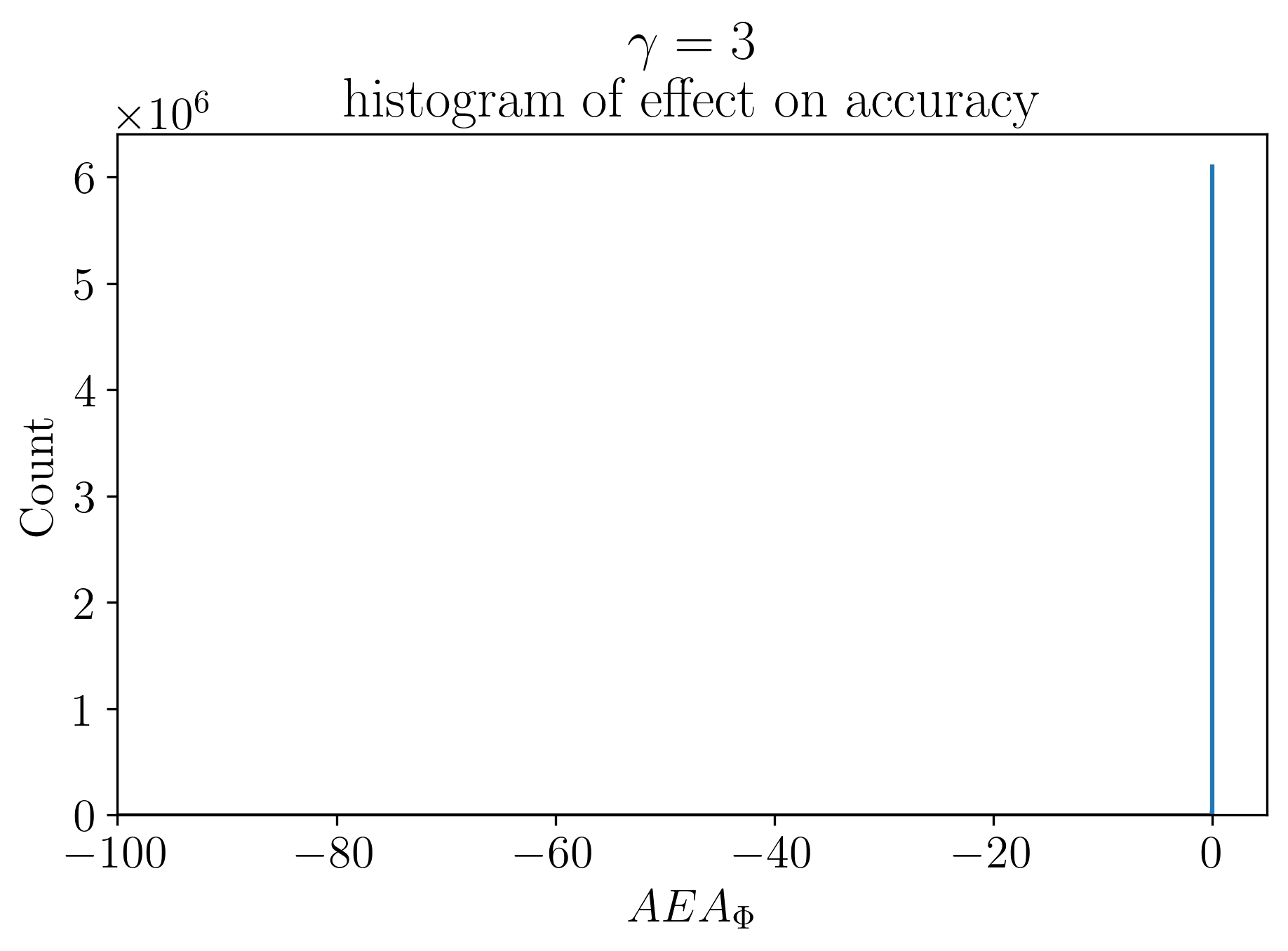}
		\caption{}
	\end{subfigure}
	\quad
	\begin{subfigure}{\spl\linewidth}
		\includegraphics[width=\linewidth]{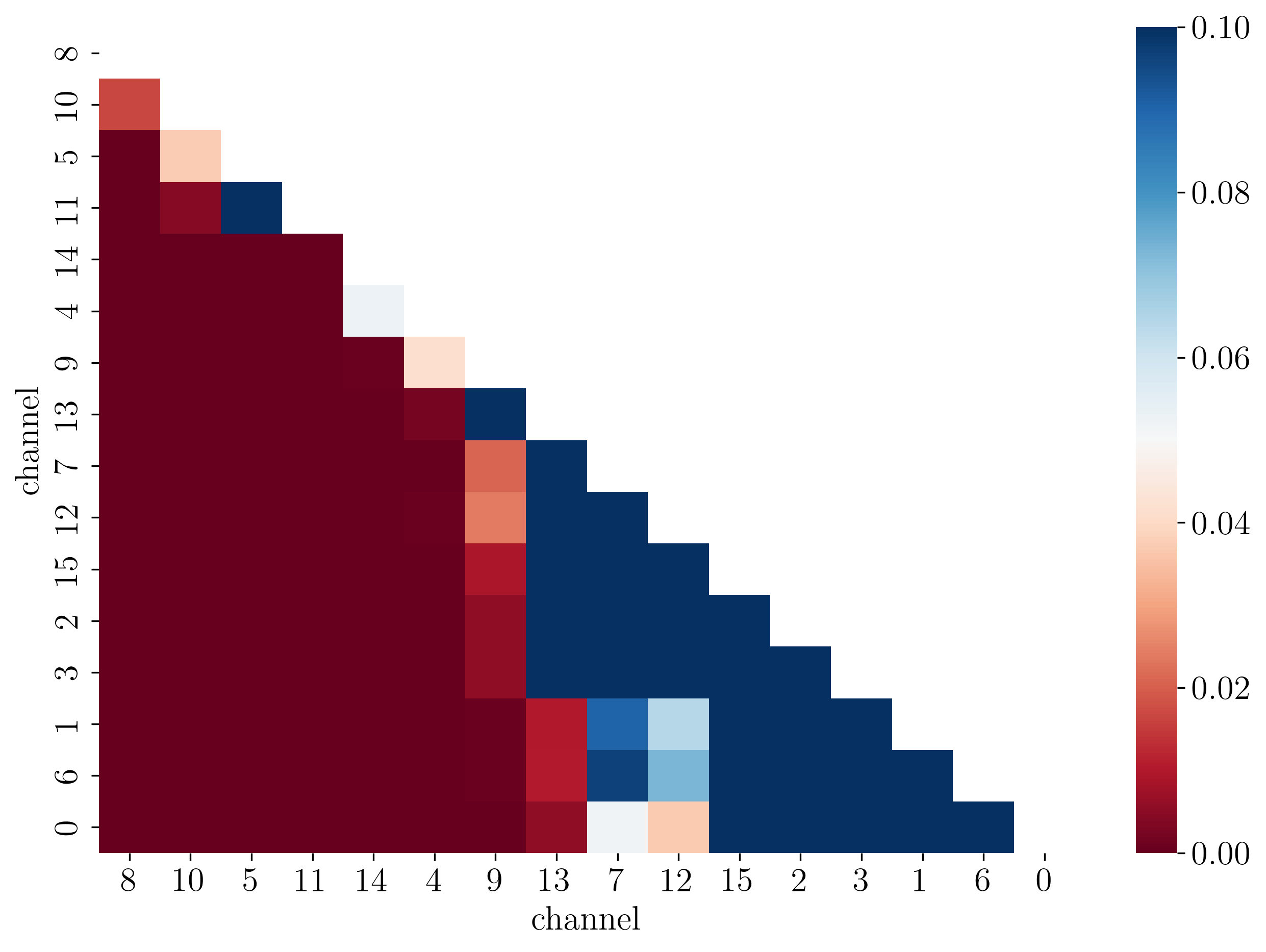}
		\caption{}
	\end{subfigure}
	\caption{Results of \textit{Adversarial Intervention} (Hop-Skip Jump) on CIFAR-10 model for $\gamma=3$. \fourfigdesc}
	\label{apx:fig:cifar:gamma3:hsj}
\end{figure*}

\begin{figure*}
	\newcommand{\spl}{0.45}
	\centering
	\begin{subfigure}{\spl\linewidth}
		\includegraphics[width=\linewidth]{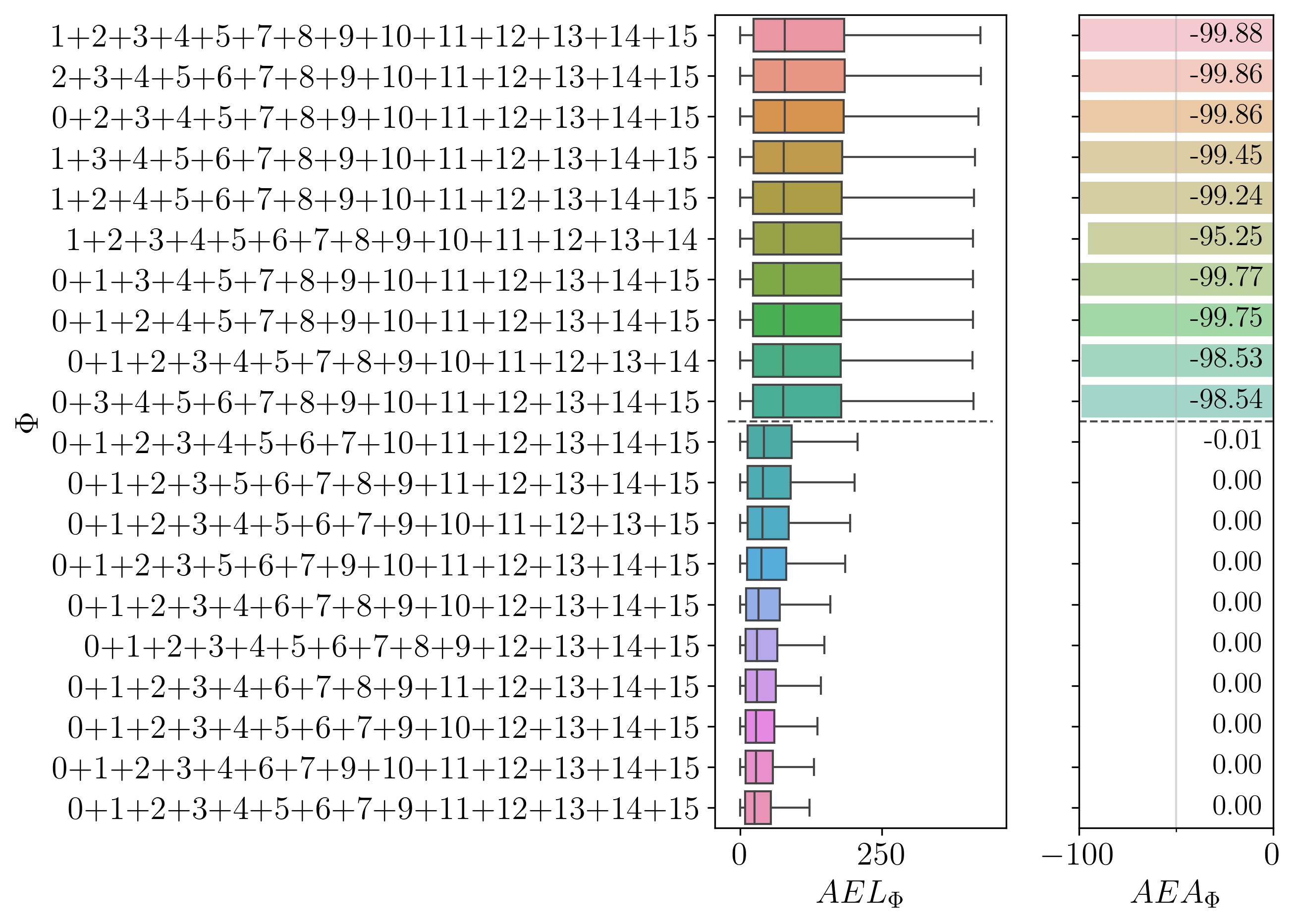}
		\caption{}
	\end{subfigure}
	\quad
	\begin{subfigure}{\spl\linewidth}
		\includegraphics[width=\linewidth]{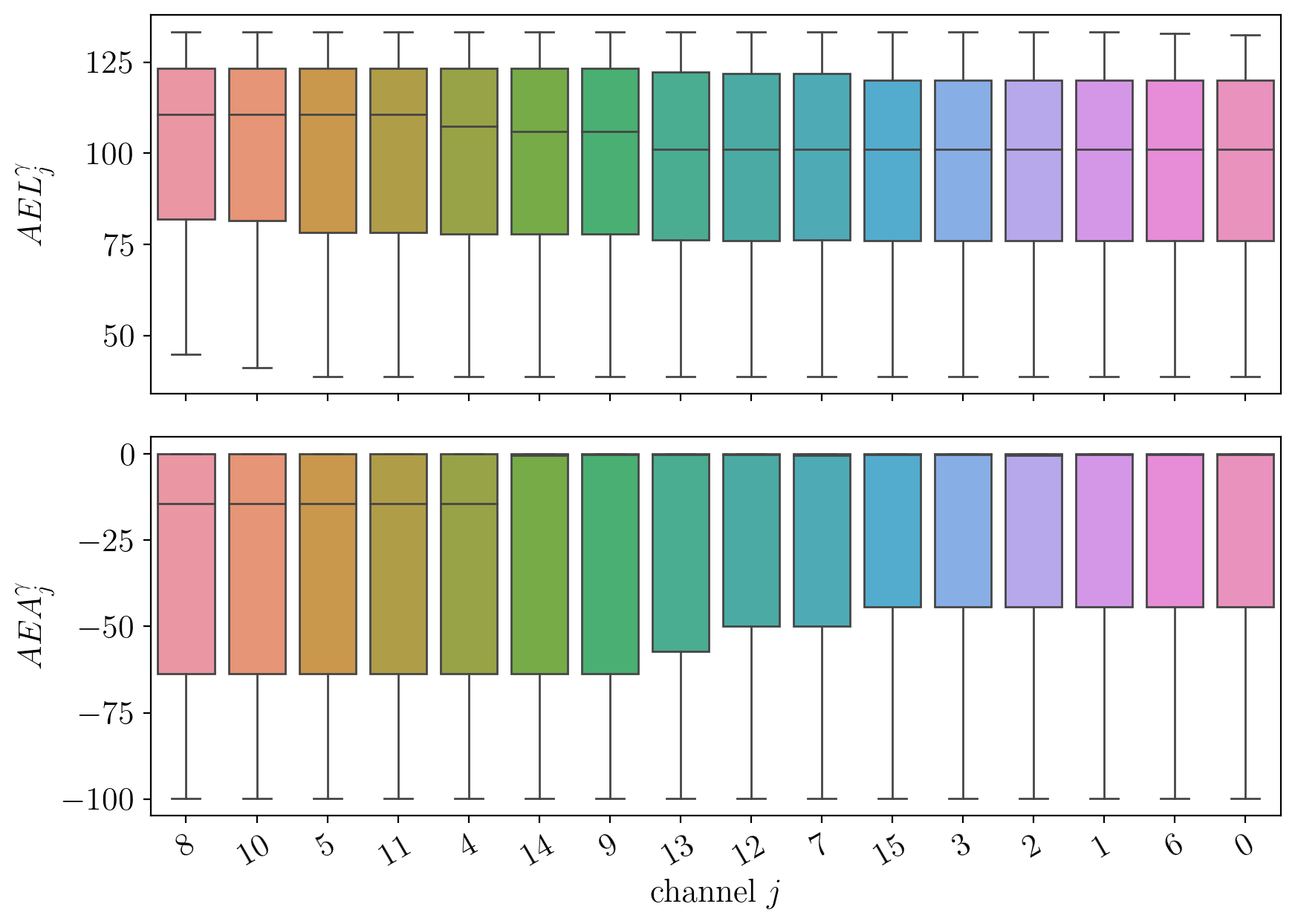}
		\caption{}
	\end{subfigure}
	\\
	\begin{subfigure}{\spl\linewidth}
		\includegraphics[width=\linewidth]{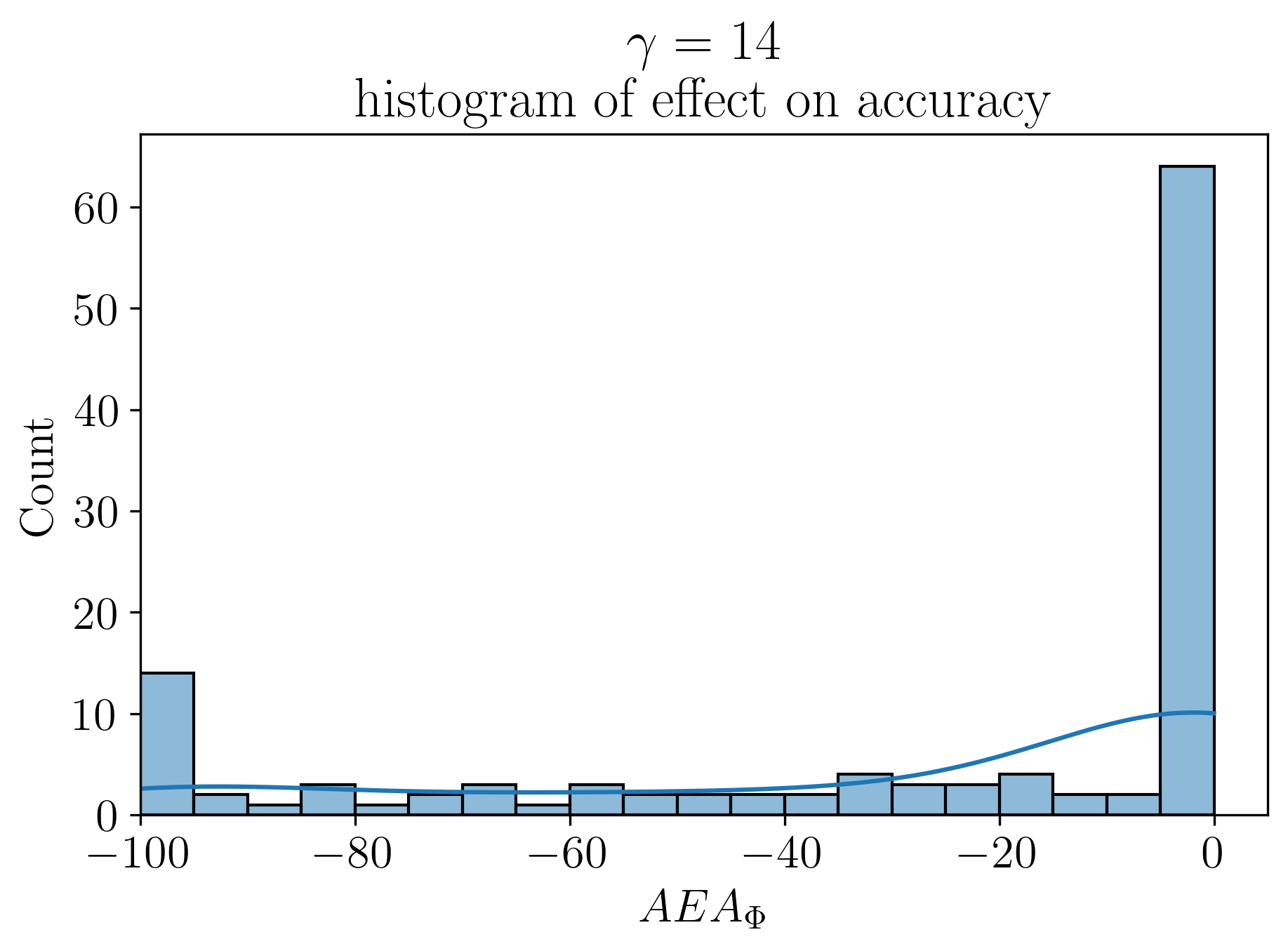}
		\caption{}
	\end{subfigure}
	\quad
	\begin{subfigure}{\spl\linewidth}
		\includegraphics[width=\linewidth]{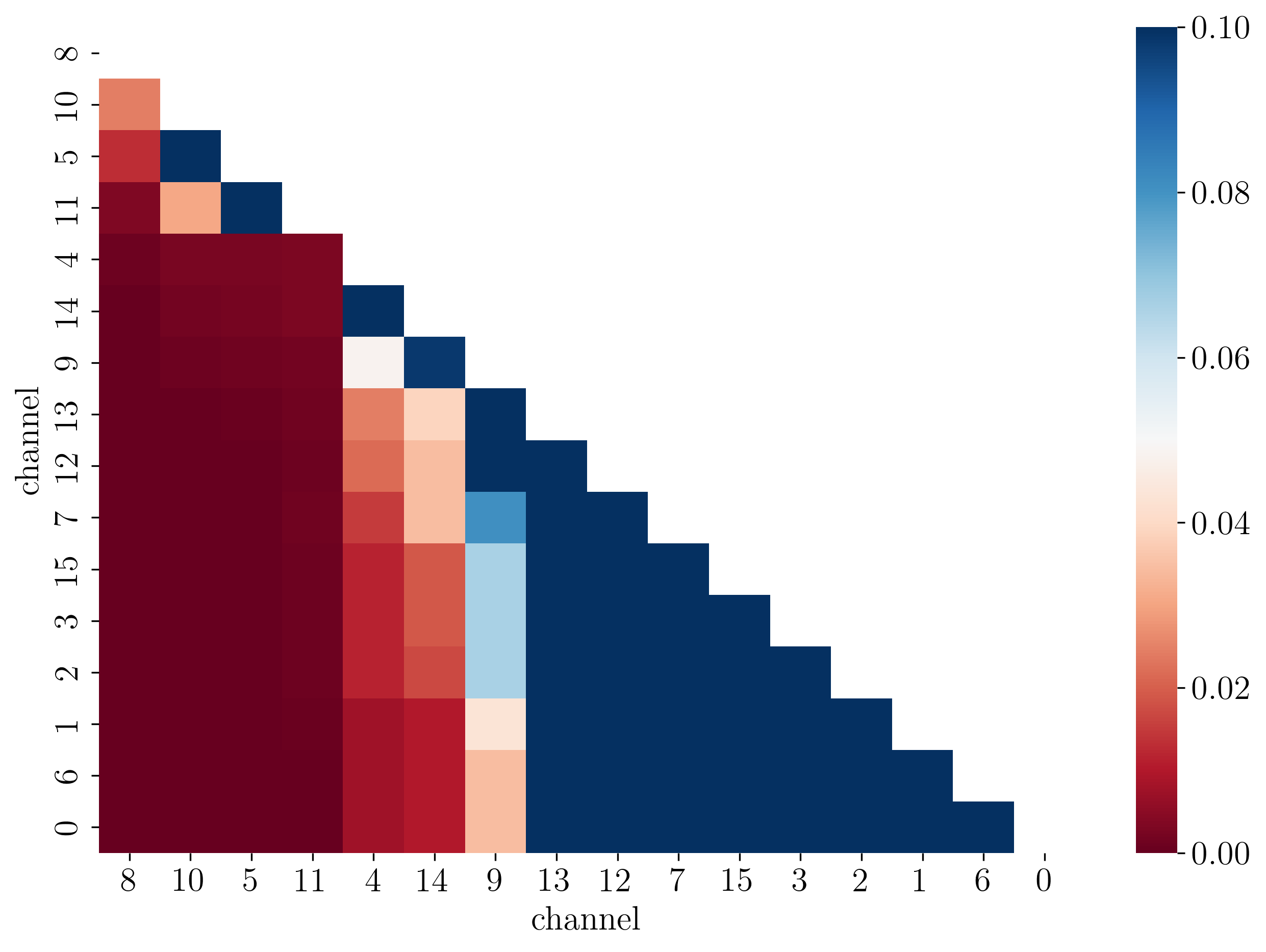}
		\caption{}
	\end{subfigure}
	\caption{Results of \textit{Adversarial Intervention} (Hop-Skip Jump) on CIFAR-10 model for $\gamma=14$. \fourfigdesc}
	\label{apx:fig:cifar:gamma14:hsj}
\end{figure*}

\begin{figure*}
	\begin{subfigure}{\linewidth}
		\includegraphics[width=\linewidth]{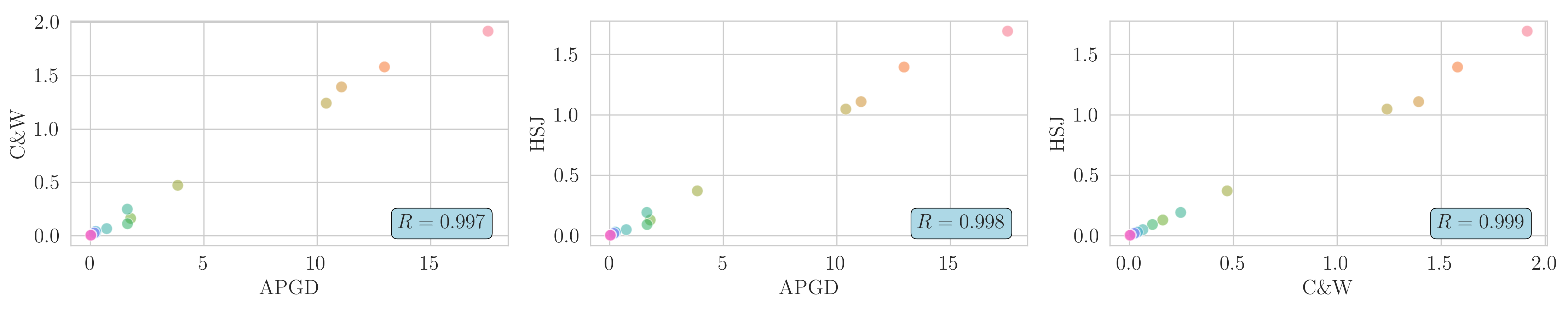}
		\caption{}
	\end{subfigure}	
	\\
	\begin{subfigure}{\linewidth}
		\includegraphics[width=\linewidth]{cifar10-compare_atks-g3.png}
		\caption{}
	\end{subfigure}	
	\\
	\begin{subfigure}{\linewidth}
		\includegraphics[width=\linewidth]{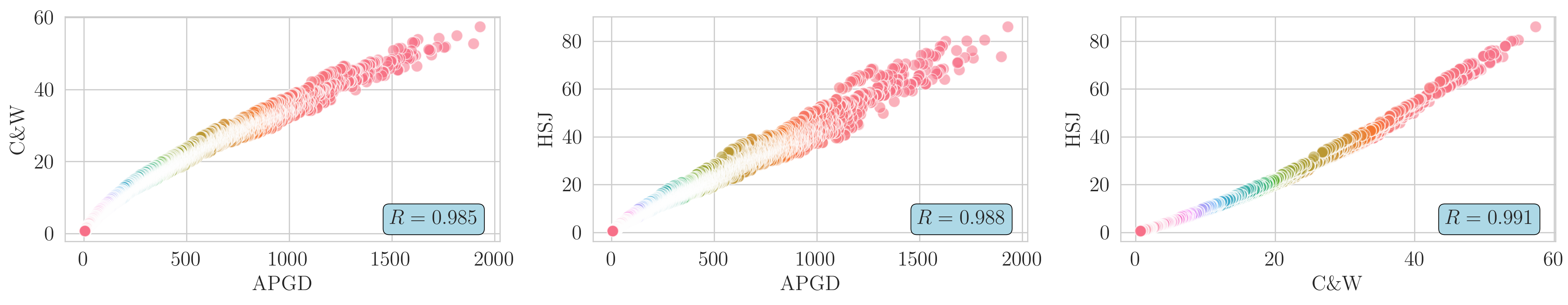}
		\caption{}
	\end{subfigure}	
	\caption{Correlation between the $AEL_\Phi$ obtained on the same CIFAR-10 model through \textit{Adversarial Intervention} with different attacks (APG: Auto-PGD; C\&W: Carlini \& Wagner; HSJ: Hop-Skip Jump). Results reported for three values of $\gamma$: (a) $\gamma=1$; (b) $\gamma=3$; (c) $\gamma=7$. Each point in the plots represent the effect caused by a channel combination $\Phi$. The plots also report the value for the $R$ correlation coefficient.}
\end{figure*}

\end{document}